\documentclass[pdflatex,sn-basic,iicol]{sn-jnl}

\usepackage{graphicx}%
\usepackage{multirow}%
\usepackage{amsmath,amssymb,amsfonts}%
\usepackage{amsthm}%
\usepackage{mathrsfs}%
\usepackage[title]{appendix}%
\usepackage{xcolor}%
\usepackage{textcomp}%
\usepackage{manyfoot}%
\usepackage{booktabs}%
\usepackage{algorithm}%
\usepackage{algorithmicx}%
\usepackage{algpseudocode}%
\usepackage{listings}%
\usepackage{colortbl}
\usepackage[dvipsnames]{xcolor}         
\usepackage{soul}
\newcommand{\ours}{FSR}

\theoremstyle{thmstyleone}%
\theoremstyle{thmstyletwo}%

\theoremstyle{thmstylethree}%

\begin{document}

\title[Focus-Scan-Refine]{Focus-Scan-Refine: From Human Visual Perception to Efficient Visual Token Pruning}


\author[1]{\fnm{Enwei} \sur{Tong}}\email{24S103434@stu.hit.edu.cn}

\author*[1]{\fnm{Yuanchao} \sur{Bai}}\email{yuanchao.bai@hit.edu.cn}

\author[2]{\fnm{Yao} \sur{Zhu}}\email{ee\_zhuy@zju.edu.cn}

\author[1]{\fnm{Junjun} \sur{Jiang}}\email{jiangjunjun@hit.edu.cn}

\author[1]{\fnm{Xianming} \sur{Liu}}\email{csxm@hit.edu.cn}

\affil*[1]{\orgdiv{Faculty of computing}, \orgname{Harbin Institute of Technology}, \orgaddress{\street{West Dazhi Street}, \city{Harbin}, \postcode{150001}, \state{Heilongjiang}, \country{China}}}

\affil[2]{\orgdiv{Chu Kochen Honors College}, \orgname{Zhejiang University}, \orgaddress{\street{Yuhangtang Road}, \city{Hangzhou}, \postcode{310058}, \state{Zhejiang}, \country{China}}}


\abstract{Vision-language models (VLMs) often generate massive visual tokens that greatly increase inference latency and memory footprint; while training-free token pruning offers a practical remedy, existing methods still struggle to balance local evidence and global context under aggressive compression.
We propose Focus-Scan-Refine (FSR), a human-inspired, plug-and-play pruning framework that mimics how humans answer visual questions: focus on key evidence, then scan globally if needed, and refine the scanned context by aggregating relevant details.
FSR first focuses on key evidence by combining visual importance with instruction relevance, avoiding the bias toward visually salient but query-irrelevant regions.
It then scans for complementary context conditioned on the focused set, selecting tokens that are most different from the focused evidence.
Finally, FSR refines the scanned context by aggregating nearby informative tokens into the scan anchors via similarity-based assignment and score-weighted merging, without increasing the token budget.
Extensive experiments across multiple VLM backbones and vision-language benchmarks show that FSR consistently improves the accuracy-efficiency trade-off over existing state-of-the-art pruning methods. The source codes can be found at \url{https://github.com/ILOT-code/FSR}}




\maketitle

\section{Introduction}
\begin{figure}[t]
    \centering
    \includegraphics[width=\linewidth]{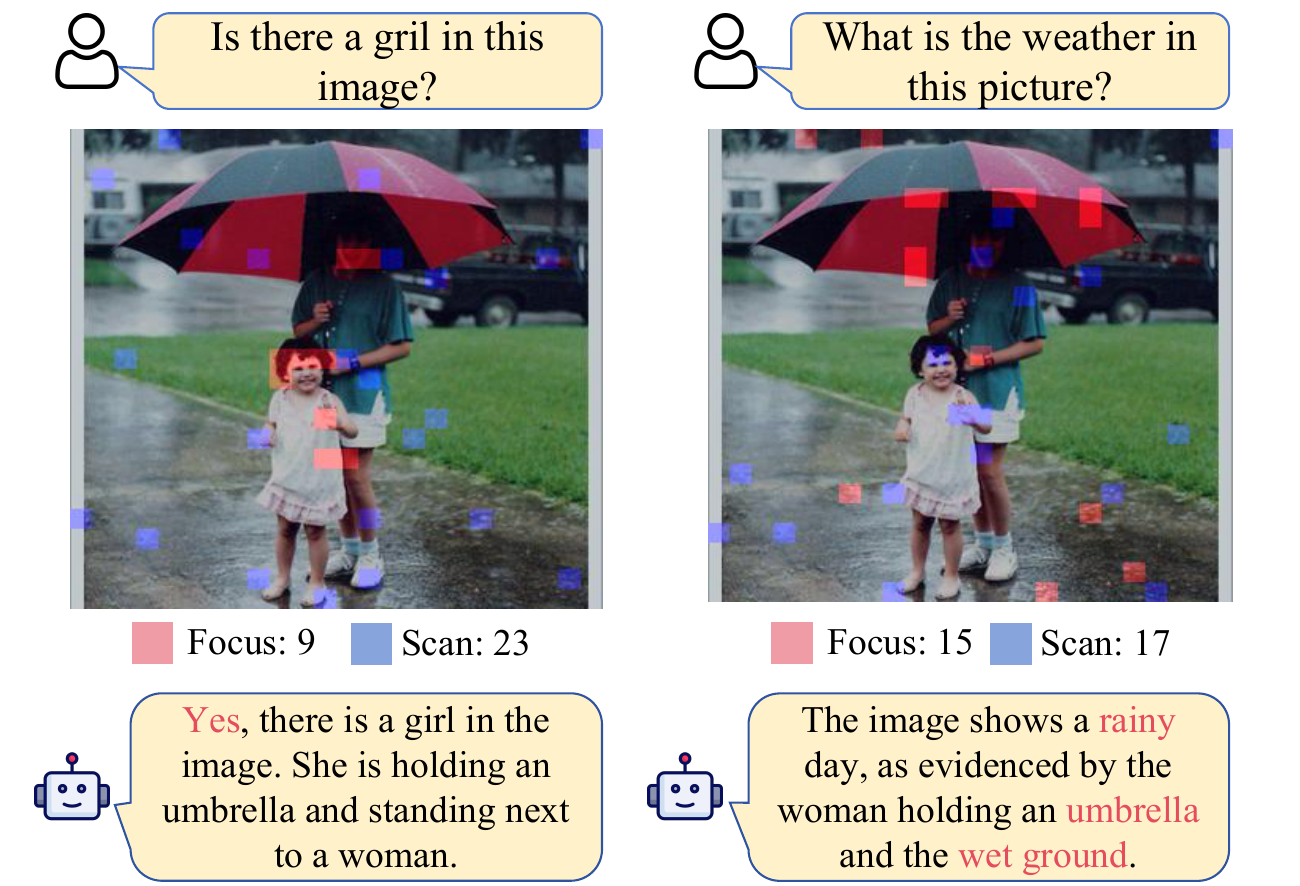}
    \caption{\textbf{Dynamic allocation of local evidence and global context.} Red tokens denote Focus (local evidence) and blue tokens denote Scan (global context). FSR dynamically reallocates the 32 token budget across tasks: for a simple existence query, it concentrates on a small local region (Focus = 9, Scan = 23), whereas for a reasoning-intensive query (weather inference), it attends to multiple cues (e.g., umbrella and wet ground), increasing local evidence coverage (Focus = 15, Scan = 17).}
    \label{fig:fsr_dynamic}
\end{figure}

With the rapid progress of large language models (LLMs)~\cite{openai2024gpt4technicalreport,touvron2023llama,jiang2023mistral7b,qwen2025qwen25technicalreport}, vision--language models (VLMs) have advanced substantially in multimodal perception and reasoning~\cite{clip,alayrac2022flamingo,li2023blip2,dai2023instructblip,liu2023llava,zhu2023minigpt4,chen2024internvl,gpt4v,gemini}.
A typical VLM encodes an image into a sequence of visual tokens, concatenates them with text tokens, and performs autoregressive decoding with an LLM.
To preserve fine details, modern VLMs increasingly adopt high-resolution encoders and tiling strategies~\cite{bai2023qwenvl,li2024llavanext,chen2024internvl}, which often produce massive visual tokens.
Since Transformer attention scales quadratically with sequence length~\cite{vaswani2017attention}, these tokens greatly increase latency and memory, becoming a key bottleneck for deployment~\cite{team2024gemma,hu2024minicpm}.
\begin{figure*}[t]
    \centering
    \includegraphics[width=\linewidth]{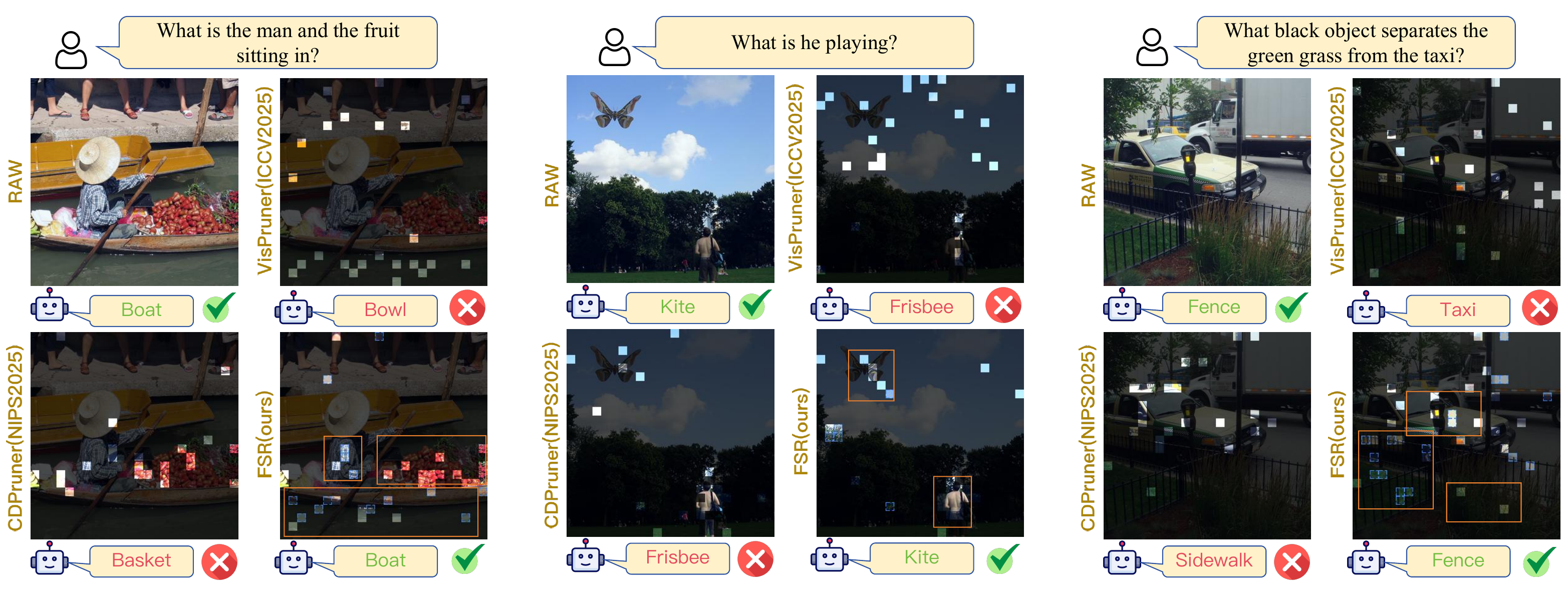}
    \caption{\textbf{Visualization-based analysis of FSR on relational visual reasoning tasks}.
    Highlighted tokens indicate the selected visual tokens, while tokens with blue borders denote those used for refinement; a fixed budget of 24 visual tokens is retained for all methods.
    In the three examples, FSR captures (i) the man, fruit, boat, as well as the surrounding water, (ii) the man and the butterfly-shaped kite he is playing with, and (iii) multiple interacting entities such as the taxi, grass, and fence.
    By contrast, VisPruner, HoloV, and CDPruner often over focus on a single local region, failing to preserve enough information to answer the question.}
    \label{fig:visual_res}
\end{figure*}
A practical remedy is training-free visual token pruning, which reduces visual tokens under a fixed budget.
Existing methods can be categorized by the signals they exploit:
\textbf{(i) Attention-based pruning} selects tokens with high cross-attention or $[\mathrm{CLS}]$-based attention, and thus tends to favor locally salient regions~\cite{fastv,prumerge};
\textbf{(ii) Similarity-based pruning} relies on inter-token similarity to encourage token diversity, and therefore tends to retain tokens that provide global scene coverage~\cite{divprune,dart};
\textbf{(iii) Joint attention–similarity-based pruning} combine both cues~\cite{visionzip,vispruner,cdpruner,holov}, but still struggle to balance local evidence and global context under high reduction ratios.

Importantly, the desired allocation between local and global tokens is task-dependent.
Tasks involving multiple objects, relations, or reasoning typically require collecting multiple local cues across different regions, while fine-grained recognition often depends on a small set of concentrated evidence.
Without a proper balance, the retained tokens are often incomplete for the target question, leaving the LLM with insufficient evidence or context for reliable reasoning.

Studies of human perception in visual question answering tasks show that humans selectively focus on task relevant regions, expand attention to scan the global context, and integrate peripheral cues via ensemble coding for a holistic representation~\cite{yarbuseye,cogitiveding,henderson2003human,alvarez2011representing}. 
Inspired by this cognitive process, we propose the Focus-Scan-Refine (FSR) pruning framework, which follows a simple three-stage design. 
\emph{(i) Focus:} we employ a dual-pathway scoring mechanism that fuses visual saliency with instruction relevance to identify critical local evidence, keeping top tokens until a cumulative information density threshold is met.
\emph{(ii) Scan:} conditioned on the focused set, we select complementary tokens that are most different from the focused evidence and diverse among themselves, ensuring the added tokens cover missing context without redundancy.
\emph{(iii) Refine:} we further strengthen global context by merging nearby informative tokens into scan anchors via similarity-based assignment and score-weighted aggregation, while keeping the token budget unchanged.

Overall, FSR dynamically adjusts the allocation between local evidence and global context according to the complexity of the input task, as illustrated in Figure~\ref{fig:fsr_dynamic}. Compared with prior methods, FSR achieves a more effective balance between local and global information, as further demonstrated in Figure~\ref{fig:visual_res}.
The main contributions are summarized as follows:
\begin{itemize}
    \item 
    We propose FSR, a human-inspired, training-free pruning framework that dynamically allocates a fixed token budget between local evidence and complementary global context, rather than relying on static local/global heuristics.

    \item 
    We introduce a comprehensive pipeline comprising a dual-pathway scoring mechanism for local evidence, a conditional sampling strategy for global context, and an aggregation module for texture refinement, ensuring efficient and non-redundant token selection.

    \item 
    Extensive experiments demonstrate that FSR consistently outperforms prior visual token pruning methods. The improvement arises from its ability to balance local evidence and global context more effectively.
\end{itemize}

\section{Related work}

The high inference cost of modern VLMs is largely driven by the massive number of visual tokens, which dominate both attention computation and KV-cache memory.
To mitigate this overhead without additional training, a growing line of work studies training-free visual token reduction.
Existing methods primarily differ in the signals used to estimate token importance.

\textbf{Attention-based Pruning.}
Attention-based pruning estimates token importance from attention statistics, either inside the LLM decoder or within the vision encoder.
On the LLM side, FastV prunes visual tokens according to cross attention scores in shallow layers~\cite{fastv}.
LLaVA-PruMerge further combines attention-based pruning with token merging to compress redundant visual tokens while preserving spatial semantics~\cite{prumerge}.
SparseVLM introduces text-guided attention scoring and token recycling to reduce information loss during progressive sparsification~\cite{sparsevlm}, while PyramidDrop (PDrop) applies layer-wise progressive dropping to better align pruning strength with model depth~\cite{pyramiddrop}.
To enhance deployment efficiency, TopV ensures FlashAttention compatibility during prefilling~\cite{ topv,dao2022flashattentionfastmemoryefficientexact} , whereas FitPrune minimizes attention-distribution divergence for budget-aware pruning~\cite{fitprune}.
On the vision-encoder side, FasterVLM and HiRED rank tokens using $[\mathrm{CLS}]$-based attention, enabling early or region aware pruning~\cite{fastervlm,hired}. SparseVILA decouples visual sparsity into query-agnostic prefill and query-aware decoding stages~\cite{sparsevila}.
Overall, while attention-based methods are effective and easy to deploy, their importance estimates can be biased toward salient regions, which may inadvertently limit the coverage of subtle yet critical global contextual information.

\textbf{Similarity-based Pruning.}
Similarity-based approaches reduce redundancy by selecting diverse visual tokens in feature space rather than relying on saliency or importance scores. These methods are motivated by the observation that attention-based criteria may not reliably reflect whether a token is redundant, and can even lead to inferior performance or incompatibility with FlashAttention.
DivPrune formulates token pruning as a max--min diversity selection problem to retain a representative and diverse subset~\cite{divprune}.
DART further prunes tokens based on duplication by retaining tokens dissimilar to a small set of pivots, enabling training-free acceleration~\cite{dart}.
However, as these methods primarily concentrate on global regions, they often overlook fine-grained local details that are essential for precise reasoning.

\textbf{Joint attention-similarity-based Pruning.} Recent methods combine multiple cues to better trade off query-critical local evidence and complementary global context.
VisionZip and VisPruner integrate attention-based importance estimation with redundancy reduction to reduce token count while maintaining coverage~\cite{visionzip,vispruner}.
CDPruner further incorporates instruction relevance and maximizes conditional diversity through a DPP-style formulation, encouraging the retained tokens to be both relevant and diverse under the prompt~\cite{cdpruner}.
HoloV promotes holistic context retention by partition-wise allocation and connectivity aware token selection, aiming to avoid over-focusing on a few highlighted regions~\cite{holov}.
Despite their effectiveness, under a fixed and limited token budget these methods can still struggle to simultaneously preserve the most query-critical local evidence and the complementary global context needed for reliable reasoning, especially when the retained tokens become extremely sparse.

\smallskip
Prior research has investigated various token pruning strategies including attention-based, similarity-based, and joint attention-similarity-based pruning. However, effectively preserving both query-critical local evidence and complementary global context remains a formidable and persistent challenge, particularly under stringent token budgets. To address this limitation, we propose FSR, a human-inspired paradigm that dynamically balances fine-grained local detail and broad global context in accordance with the intrinsic complexity of the input.

\section{Proposed Method}
\label{sec:method}

\begin{figure}[t]
    \centering
    \includegraphics[width=\linewidth]{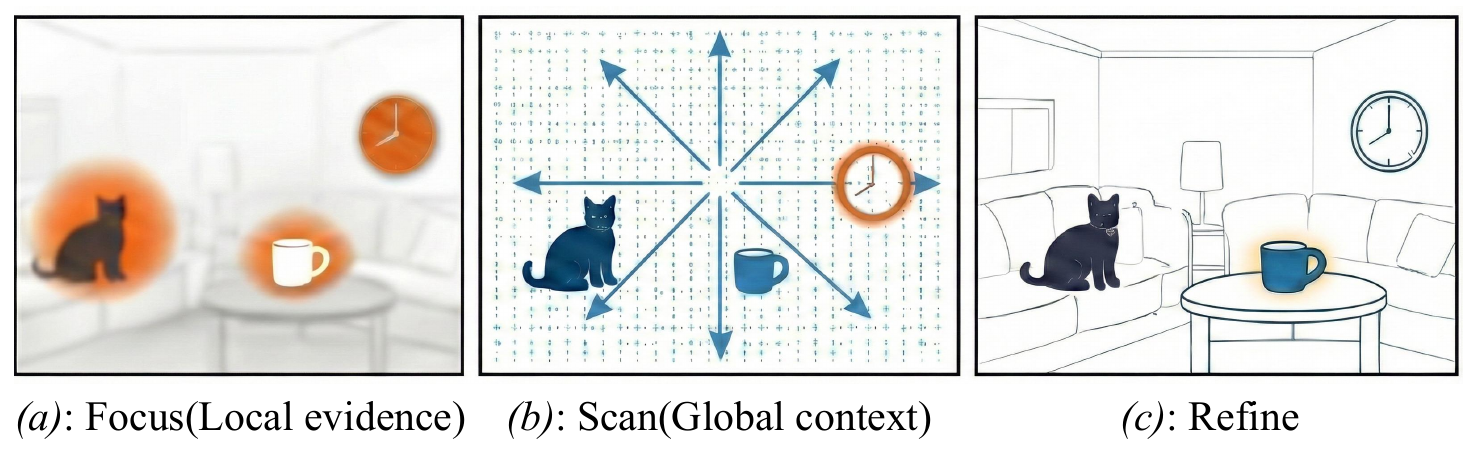}
    \caption{\textbf{Human Visual Perceptual Strategy under Limited Attention.}
    (a) Constrained by finite attentional capacity, humans prioritize local regions that are most relevant to the query.
    (b) To acquire complementary information, humans expand their field of view to scan the global layout and background context.
    (c) The brain utilizes ensemble coding to aggregate peripheral signals into summary statistics, forming a robust global representation.}
\label{fig:humanvis}
\end{figure}

\begin{figure*}[!t]
\centering
\includegraphics[width=\linewidth]{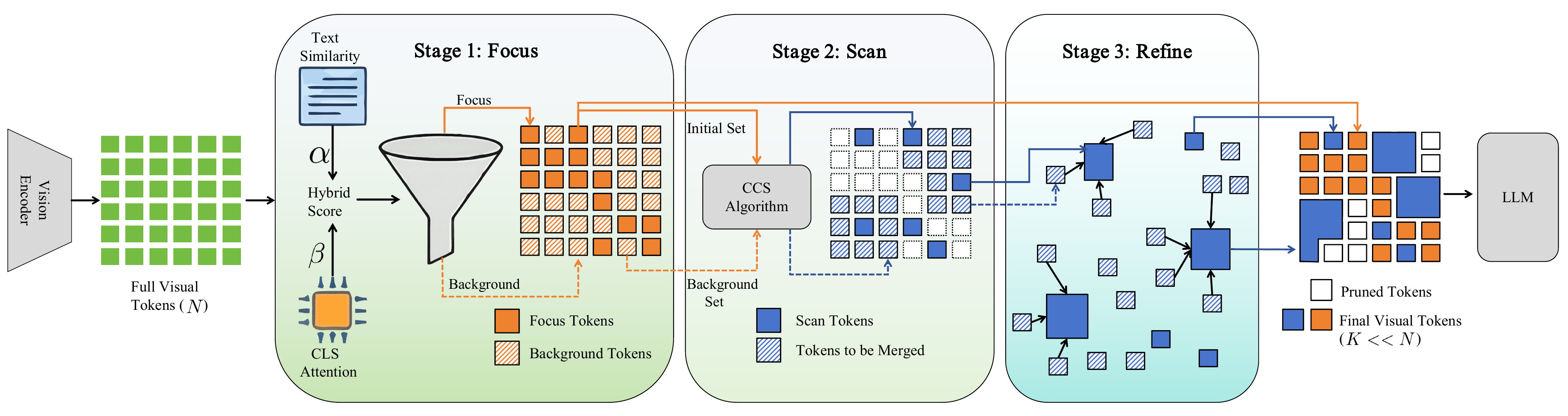}
\caption{\textbf{Overview of the FSR framework.} 
Given input visual tokens and a query, FSR progressively compresses information into a fixed budget $K$: 
(1) Focus: Identifies critical local evidence ($\mathcal{F}$) via a dual-pathway scoring mechanism fusing visual saliency and instruction relevance. 
(2) Scan: Captures complementary global context ($\mathcal{S}$) using the Conditional Context Sampling (CCS) algorithm to maximize information gain. 
(3) Refine: Enriches the sparse context anchors by aggregating relevant discarded details via weighted merging, ensuring a holistic representation for the LLM.}
\label{fig:overview}
\end{figure*}

\subsection{Inspiration from the Human Visual Perception}
Our methodology is inspired by how the human visual system allocates perceptual resources under limited attention. Cognitive science research indicates that when answering visual questions, humans do not process the entire scene with equal fidelity; instead, they prioritize extracting information from local regions highly relevant to the query~\cite{yarbuseye, cogitiveding}. 
Reliance on local cues alone is often insufficient for complex tasks; when initial local evidence fails to yield a confident answer, humans scan the global context to find more cues~\cite{henderson2003human, wolfe2017five}. Subsequently, rather than discarding the remaining peripheral information, the brain utilizes ensemble coding to aggregate it into summary statistics, ensuring a complete yet efficient scene representation~\cite{alvarez2011representing}. Figure~\ref{fig:humanvis} provides a high-level illustration of this general organization of human visual processing. 

Inspired by this perceptual strategy of progressively allocating attention from local evidence to global context, we propose the FSR framework (see Figure~\ref{fig:overview} for an overview) to simulate this progressive process. 
To mathematically instantiate this progressive process, we model the task as identifying an optimal subset of tokens within an explicitly constrained budget.

Given an input image, a vision encoder outputs a sequence of visual tokens $\mathbf{V}=\{\mathbf{v}_i\}_{i=1}^{N}$ where $\mathbf{v}_i\in\mathbb{R}^{d}$. 
Given a query $\mathbf{q}$ and a token budget $K$ ($K\ll N$), our objective is to identify a compressed subset $\widetilde{\mathbf{V}} \subset \mathbf{V}$ with $|\widetilde{\mathbf{V}}|=K$.
Unlike static pruning, FSR dynamically constructs $\widetilde{\mathbf{V}}$ by first locking onto key local evidence (Focus) and then expanding the field of view (Scan \& Refine) to get more contextual information.

\subsection{Stage I: Focus on local evidence}
\label{subsec:focus}
The \textbf{Focus} stage aims to identify and retain the most critical local visual evidence, mimicking the focus mechanism in human visual perception. To avoid the potential bias of relying solely on a single signal, we employ a dual-pathway scoring mechanism fusing both visual saliency and instruction relevance, ensuring that the selected tokens are not only visually salient but also semantically aligned with the user's instruction.

We first identify inherently salient regions (e.g., foreground objects) using the attention map from the vision encoder. Denote by $\mathbf{A}\in\mathbb{R}^{H\times (N+1)\times (N+1)}$, the attention map from the $[\mathrm{CLS}]$ token to other tokens in a selected layer. The saliency score $s_i$ for the $i$-th token is computed as:
\begin{equation}
s_i=\frac{1}{H}\sum_{h=1}^{H}\mathbf{A}_h[\mathrm{CLS},i]
\end{equation}
To ensure that the selected tokens are relevant to the user's instruction, we compute the semantic similarity between visual tokens and the text instruction~\cite{cdpruner}. We encode the textual query $\mathbf{q}$ into an embedding $\mathbf{t}$ using the pre-trained CLIP text encoder. The relevance score $r_i$ is defined as the cosine similarity:
\begin{equation}
\begin{split}
r_i &= \cos(\bar{\mathbf{v}}_i,\bar{\mathbf{t}}), \\
\text{where } \bar{\mathbf{v}}_i = &{\mathbf{v}_i}/{\|\mathbf{v}_i\|_2}, \: \bar{\mathbf{t}}=\mathbf{t}/{\|\mathbf{t}\|_2}
\end{split}
\end{equation}
We further normalize both scores to $[0,1]$ (denoted by the hat notation $\hat{\cdot}$) and compute a fused priority score $\phi_i$ to generate a unified priority map:
\begin{equation}
\phi_i=\hat{r}_i^{\alpha}\hat{s}_i^{\beta}
\label{eq:fused_score}
\end{equation}
where $\alpha$ and $\beta$ control the trade-off between relevance and saliency. 
Tokens are then sorted by $\phi$ in descending order, denoted by the permutation $\pi$. To determine the dynamic budget $K_{\mathrm{F}}$, we select the minimum number of tokens required to preserve a ratio $\rho$ (default 0.9) of the total information mass $Z=\sum_{i=1}^{N}\phi_{i}$:
\begin{equation}
K_{\mathrm{F}} = \min \left\{ k \mid \sum_{j=1}^{k}\phi_{\pi(j)} \ge \rho Z \right\}
\label{eq:focus_size}
\end{equation}
The resulting set $\mathcal{F}=\{\pi(1),\ldots,\pi(K_{\mathrm{F}})\}$ constitutes the local evidence.

\subsection{Stage II: Scan for global context}
\label{subsec:scan}
\subsubsection{Conditional Context Sampling}
Relying solely on local evidence $\mathcal{F}$ often results in missing critical background information required for holistic reasoning. The \textbf{Scan} stage addresses this by expanding the attentional window to capture broader global context when local information is insufficient.

We introduce a Conditional Context Sampling (CCS) algorithm to select $K_{\mathrm{S}}=K-K_{\mathrm{F}}$ supplementary anchors. To maximize information gain, these anchors must be complementary to the focused set $\mathcal{F}$ and diverse among themselves. 
Specifically, we initialize the available anchor set as $\mathcal{A}=\mathcal{F}$. In each iteration, we identify the token $i^\star$ that is maximally different from the current anchor set $\mathcal{A}$ in the feature space:
\begin{equation}
\begin{aligned}
\Delta(i,\mathcal{A})&=\min_{j\in\mathcal{A}}\Bigl(1-\cos(\bar{\mathbf{v}}_i,\bar{\mathbf{v}}_j)\Bigr), \\
i^\star&=\arg\max_{i\notin\mathcal{A}}\Delta(i,\mathcal{A})
\label{eq:scan_select}
\end{aligned}
\end{equation}
We update $\mathcal{A} \leftarrow \mathcal{A} \cup \{i^\star\}$ and repeat this process for $K_{\mathrm{S}}$ iterations. This strategy ensures that the newly captured tokens are different from the salient objects and minimizes redundancy, thereby optimizing the utility of the token budget. Finally, the specific set of scanned context tokens is obtained as $\mathcal{S} = \mathcal{A} \setminus \mathcal{F}$.

\subsubsection{Theoretical Coverage Guarantee}
While the CCS strategy is greedy, it admits a formal coverage guarantee, ensuring that the selected context tokens provide bounded approximation to the optimal global coverage.

The CCS procedure in Eq.~\eqref{eq:scan_select} can be viewed as a variant of Farthest Point Sampling~\cite{Gonzalez1985ClusteringTM} in the feature space, where the focus set $\mathcal{F}$ is treated as a fixed set of initial centers.
Let $V$ denote the set of all visual tokens, equipped with the distance metric $d(x,y)=1-\cos(x,y)$.
Given a total budget $K$ and the fixed focus set $\mathcal{F}$, we define the optimal conditional covering radius as
\begin{equation}
R_{\mathrm{opt}}(\mathcal{F})=\min_{S': |S'|=K-|\mathcal{F}|}\max_{v\in V} d\bigl(v,\mathcal{F}\cup S'\bigr)
\end{equation}
This quantity represents the minimum achievable worst-case distance when extending $\mathcal{F}$ with $K_{\mathrm{S}}$ additional tokens.
By classical results on greedy $k$-center clustering with fixed centers~\cite{hochbaum1985best}, the token set $\mathcal{K}=\mathcal{F}\cup\mathcal{S}$ selected by CCS satisfies:
\begin{equation}
\max_{v\in V}\min_{u\in\mathcal{K}} d(v,u)\le 2\,R_{\mathrm{opt}}(\mathcal{F})
\end{equation}
which bounds the information loss incurred by pruning.
This guarantee implies that CCS attains a near globally optimal solution, ensuring that every unselected token lies within a bounded distance of the selected token set.

\subsection{Stage III: Refine by aggregation}
\label{subsec:refine}
Directly discarding the unselected tokens $\mathcal{D} = \mathbf{V} \setminus (\mathcal{F} \cup \mathcal{S})$ leads to a loss of fine-grained background details. The \textbf{Refine} stage addresses this by aggregating information from the discarded set $\mathcal{D}$ into the selected context anchors.

Crucially, to preserve the high fidelity of the salient objects, we keep the focus set $\mathcal{F}$ unchanged. We treat only the global context tokens $\mathcal{S}$ as semantic anchors for aggregation.
First, for each discarded token $i \in \mathcal{D}$, we identify its semantically nearest anchor $j^\star$ within the scan set $\mathcal{S}$ and compute their similarity:
\begin{equation}
j^\star(i) = \arg\max_{j\in\mathcal{S}} \cos(\bar{\mathbf{v}}_i,\bar{\mathbf{v}}_j)
\label{eq:nearest_anchor}
\end{equation}
To mitigate noise and prevent over-smoothing, we do not aggregate all discarded tokens. Instead, we select the top-$M$ tokens from the discarded set $\mathcal{D}$ that possess the highest similarity scores to their assigned anchors. The total aggregation budget is dynamically determined by the size of the scan set as $M = \kappa |\mathcal{S}|$, where $\kappa$ is a hyperparameter set to 1 by default.
Let $\mathcal{D}_{\text{top}}$ denote this subset of highly relevant discarded tokens. We update the anchors by absorbing information only from $\mathcal{D}_{\text{top}}$. For each $i \in \mathcal{D}_{\text{top}}$, its feature is aggregated into its nearest anchor $\mathbf{v}_{j^\star}$ weighted by its priority score $\phi_i$ (from Eq.~\eqref{eq:fused_score}), as defined below:
\begin{equation}
\begin{aligned}
\mathbf{v}_{j^\star} &\leftarrow \frac{w_{j^\star}\mathbf{v}_{j^\star}+w_i\,\mathbf{v}_i}{w_{j^\star}+w_i}, \\
&w_{j^\star} \leftarrow w_{j^\star}+w_i
\end{aligned}
\label{eq:refine_merge}
\end{equation}
where weights are initialized as $w_j = \phi_j$. 
This step enables the sparse context anchors to capture the essential texture and semantics of their neighborhoods. The final compressed token set is the union of the intact focus tokens and the refined context tokens: $\widetilde{\mathbf{V}} = \mathcal{F} \cup \mathcal{S}$, which contains exactly $K_{\mathrm{F}}+K_{\mathrm{S}}=K$ tokens.

\section{Experiment}
\begin{table*}[t]
  \centering
  \caption{Performance comparison of different pruning methods on LLaVA-1.5-7B. \textbf{Avg.} represents the average relative performance maintained across all tested benchmarks compared to the unpruned baseline. The best results are highlighted in \textbf{bold}.}
  \label{tab:llava-1.5}
  \resizebox{\linewidth}{!}{
    \begin{tabular}{l|ccccccccc|c}
    \toprule
    \textbf{Method} & \textbf{$\text{VQA}^\text{V2}$} & \textbf{GQA} & \textbf{$\text{SQA}^\text{IMG}$} & \textbf{$\text{VQA}^\text{Text}$} & \textbf{POPE} & \textbf{MME} & \textbf{$\text{MMB}^\text{EN}$} & \textbf{$\text{MMB}^\text{CN}$} & \textbf{MMVet} & \textbf{Avg.} \\
    \rowcolor{lightgray!75}\multicolumn{11}{c}{\textit{Upper Bound, All 576 Tokens} ($\mathbf{100\%}$)} \\
    \rowcolor{lightgray!25} LLaVA-1.5-7B & 78.5 & 61.9 & 69.5 &	58.2 &	85.9 &	1862 &64.6 &	58.1 &	31.7& 100\% \\
    \rowcolor{lightgray!75}\multicolumn{11}{c}{\textit{Retain 192 Tokens} \textcolor{Green}{($\downarrow\mathbf{66.7\%}$)}} \\
    FastV{\small\texttt{(ECCV24)}} & 67.1 & 52.7  & 65.7 & 52.5 & 64.8 & 1612 & 61.2 & 57.0 & -  & 88.0\% \\
    SparseVLM{\small\texttt{(ICML2025)}} & 75.6 & 57.6  & 67.5 & 56.1 & 83.6 & 1721 & 62.5 & 53.7 & -  & 95.2\% \\
    DART{\small\texttt{(EMNLP2025)}} & 76.7 & 58.9  & 68.2 & 57.4 & 82.8 & 1856 & 63.6 & 57.0 & -  & 97.8\% \\
    HoloV{\small\texttt{(NIPS2025)}} & 76.4 & 58.7  & 67.2 & 55.8 & 85.0 & 1759 & 62.6 & 55.3 & 31.5  & 96.5\% \\
    VisPruner{\small\texttt{(ICCV2025)}} & 76.9 & 59.5  & 68.5 & 57.4 & 85.8 & 1780 & 63.1 & 57.0 & 33.3  & 98.2\% \\
    CDPruner{\small\texttt{(NIPS2025)}} & 77.2 & 60.3  & 68.8 & 57.3 & 87.3 & 1784 & 63.1 & 55.6 & 33.9  & 98.5\% \\
    \ours  & 77.4 & 60.2  & 69.1 & 57.6 & 87.1 & 1803 & 64.0 & 56.5 & 33.9  & \textbf{99.1\%} \\

    \rowcolor{lightgray!75}\multicolumn{11}{c}{\textit{Retain 128 Tokens} \textcolor{Green}{($\downarrow\mathbf{77.8\%}$)}} \\
    FastV{\small\texttt{(ECCV24)}} & 71.0 & 54.0 & 69.2 & 56.4 & 68.2 & 1490 & 63.0 & 55.9 & 27.0 &  89.6\% \\
    SparseVLM{\small\texttt{(ICML25)}} & 75.1 & 57.3  & 69.0 & 56.3 & 83.1 & 1696 & 62.6 & 56.9 & 29.7 &  95.6\% \\
    DART{\small\texttt{(2025.02)}} & 74.7 & 57.9  & 69.1 & 56.3 & 80.4 & 1701 & 60.7 & 57.3 & 30.9 &   95.2\% \\
    VisionZip{\small\texttt{(CVPR25)}} & 75.6 & 57.6  & 68.7 & 56.9 & 83.3 & 1721 & 62.1 & 57.0 & 31.6 &  96.2\% \\
    DivPrune{\small\texttt{(CVPR25)}} & 76.0 & 59.4  & 68.6 & 55.9 & 87.0 & 1698 & 61.5 & 54.8 & 30.6 &  96.2\% \\
    HoloV{\small\texttt{(NIPS2025)}} & 75.4 & 57.5  & 68.9 & 55.7 & 82.2 & 1766 & 62.4 & 56.8 & 31.2  & 96.1\% \\
    VisPruner{\small\texttt{(ICCV2025)}} & 75.7 & 58.5  & 69.0 & 57.0 & 84.5 & 1747 & 61.8 & 56.5 & 31.2  & 96.7\%\\
    CDPruner{\small\texttt{(NIPS2025)}} & 76.5 & 59.8  & 69.0 & 56.2 & 87.6 & 1775 & 63.1 & 55.1 & 30.9 & 97.6\% \\
    \ours  & 76.7 & 59.7  & 68.8 & 57.0 & 86.5 & 1769 & 63.2 & 55.8 & 34.9 & \textbf{98.3\%}\\
    \rowcolor{lightgray!75}\multicolumn{11}{c}{\textit{Retain 64 Tokens} \textcolor{Green}{($\downarrow\mathbf{88.9\%}$)}} \\
    FastV{\small\texttt{(ECCV24)}} & 55.9 & 46.0  & 70.1 & 51.6 & 35.5 &  1256 & 50.1 & 42.1 & 18.9 &   72.0\% \\
    SparseVLM{\small\texttt{(ICML25)}} & 66.9 & 52.0  & 69.2 & 52.1 & 69.7 & 1505 & 58.3 & 49.6 & 24.4 &   86.0\% \\
    DART{\small\texttt{(2025.02)}} & 71.3 & 54.7 & 69.3 & 54.7 & 73.8 & 1650 & 59.5 & 54.0 & 26.5 &   90.8\% \\
    VisionZip{\small\texttt{(CVPR25)}} & 72.4 & 55.1  & 69.0 & 55.5 & 77.0 & 1673 & 60.1 & 55.4 & 29.4 &  92.7\% \\
    DivPrune{\small\texttt{(CVPR25)}} & 74.1 & 57.5  & 68.0 & 54.5 & 85.5 & 1617 & 60.1 & 52.3 & 28.1 &  93.3\% \\
    HoloV{\small\texttt{(NIPS2025)}} & 72.6 & 55.1 & 68.7 & 54.8 & 76.8 & 1699 & 60.0 & 55.8 & 30.2 & 92.9\% \\
    VisPruner{\small\texttt{(ICCV2025)}} & 72.8 & 55.8  & 68.8 & 55.8 & 80.9 & 1661 & 59.4 & 54.6 & 31.4  & 93.5\% \\
    CDPruner{\small\texttt{(NIPS2025)}} & 75.4 & 58.6  & 68.1 & 55.1 & 87.5 & 1710 & 60.8 & 55.3 & 29.6 &  95.7\% \\
    \ours  & 75.4 & 58.2  & 69.3 & 55.7 & 85.7 & 1701 & 61.9 & 53.9 & 32.6 & \textbf{96.1\%} \\
    \bottomrule
    \end{tabular}
  }
  \vspace{-4mm}
\end{table*}

\begin{table*}[t]
  \centering
  \caption{Performance comparison of different pruning methods on LLaVA-NeXT-7B. \textbf{Avg.} represents the average relative performance maintained across all tested benchmarks compared to the unpruned baseline. The best results are highlighted in \textbf{bold}.}
  \label{tab:llava-next}
  \resizebox{\linewidth}{!}{
    \begin{tabular}{l|ccccccccc|cc}
    \toprule
    \textbf{Method} & \textbf{$\text{VQA}^\text{V2}$} & \textbf{GQA}& \textbf{$\text{SQA}^\text{IMG}$} & \textbf{$\text{VQA}^\text{Text}$} & \textbf{POPE} & \textbf{MME} & \textbf{$\text{MMB}^\text{EN}$} & \textbf{$\text{MMB}^\text{CN}$} & \textbf{MMVet} & \textbf{Avg.} \\
    \rowcolor{lightgray!75}\multicolumn{11}{c}{\textit{Upper Bound, All 2880 Tokens} ($\mathbf{100\%}$)} \\
    \rowcolor{lightgray!25} LLaVA-NeXT-7B & 81.3 & 62.5 & 67.6 & 60.3 & 86.8 & 1883 & 65.9 & 57.4 & 39.2 & 100.0\% \\
    \rowcolor{lightgray!75}\multicolumn{11}{c}{\textit{Retain 960 Tokens} \textcolor{Green}{($\downarrow\mathbf{66.7\%}$)}} \\
    HoloV{\small\texttt{(NIPS2025)}}& 78.9 & 61.3 & 66.2 & 57.4 & 86.9 & 1713 & 50.9 & 42.3 & 34.4  & 91.7\% \\
    VisPruner{\small\texttt{(ICCV2025)}}& 80.0 & 62.1 & 68.2 & 60.2 & 87.1 & 1807 & 65.8 & 58.2 & 38.5 & 99.2\% \\
    CDPruner{\small\texttt{(NIPS2025)}} & 80.5 & 62.7 & 68.5 & 59.1 & 87.1 & 1799 & 66.9 & 57.6 & 39.0 & 99.4\% \\
    \ours & 80.5 & 62.6 & 68.5 & 60.3 & 87.1 & 1806 & 66.9 & 58.3 & 41.1  & \textbf{100.0\%} \\
    \rowcolor{lightgray!75}\multicolumn{11}{c}{\textit{Retain 640 Tokens} \textcolor{Green}{($\downarrow\mathbf{77.8\%}$)}} \\
    FastV{\small\texttt{(ECCV24)}} & 77.0 & 58.9  & 67.4 & 58.1 & 79.5 & 1667 & 63.1 & 53.5 & 39.5 & 94.4\% \\
    DivPrune{\small\texttt{(CVPR25)}} & 79.3 & 61.9 & 67.8 & 57.0 & 86.9 & 1734 & 65.8 & 57.3 & 38.0 & 97.7\% \\
    HoloV{\small\texttt{(NIPS2025)}}& 79.3 & 61.2 & 63.8 & 57.6 & 86.2 & 1768 & 64.3 & 56.7 & 38.9 & 97.0\% \\
    VisPruner{\small\texttt{(ICCV2025)}}& 78.8 & 61.1 & 68.3 & 60.0 & 85.9 & 1828 & 64.9 & 57.3 & 38.5 & 98.5\% \\
    CDPruner{\small\texttt{(NIPS2025)}}& 79.8 & 62.6 & 68.0 & 58.5 & 87.3 & 1800 & 66.2 & 57.6 & 41.0 & 99.3\% \\
    \ours & 79.7 & 62.3 & 67.9 & 60.0 & 87.0 & 1833 & 66.3 & 57.9 & 41.9 & \textbf{99.9\%} \\
    \rowcolor{lightgray!75}\multicolumn{11}{c}{\textit{Retain 320 Tokens} \textcolor{Green}{($\downarrow\mathbf{88.9\%}$)}} \\
    FastV{\small\texttt{(ECCV24)}} & 61.5 & 49.8  & 66.6 & 52.2 & 49.5 & 1302 & 53.4 & 42.5 & 20.0 & 74.9\% \\
    DivPrune{\small\texttt{(CVPR25)}} & 77.2 & 61.1 & 67.7 & 56.2 & 84.7 & 1687 & 63.9 & 55.7 & 34.8 & 95.2\% \\
    HoloV{\small\texttt{(NIPS2025)}}& 77.2 & 59.8 & 66.2 & 57.0 & 83.4 & 1753 & 65.5 & 57.0 & 36.5 & 96.0\% \\
    VisPruner{\small\texttt{(ICCV2025)}}& 75.9 & 58.7 & 68.6 & 59.0 & 81.4 & 1753 & 63.8 & 55.8 & 36.3 & 95.4\% \\
    CDPruner{\small\texttt{(NIPS2025)}}& 78.4 & 61.4 & 67.7 & 57.4 & 87.3 & 1773 & 65.4 & 55.6 & 36.7 & 97.3\% \\
    \ours & 77.9 & 60.9 & 68.1 & 58.1 & 86.1 & 1783 & 64.9 & 56.1 & 39.3 & \textbf{97.6\%} \\
    \bottomrule
    \end{tabular}
  }
\end{table*}

\subsection{Experimental setup}
In this section, we describe the experimental configurations used to evaluate the proposed FSR framework, including the model architectures, implementation details, and benchmarks.

\textbf{Model architectures.}
We evaluate FSR on a diverse set of VLMs covering both image and video modalities. 
For static image understanding, we use the LLaVA series (LLaVA-1.5-7B/13B, LLaVA-NeXT-7B/13B) and Qwen2.5-VL-7B. 
For video understanding, we extend our evaluation to LLaVA-Video-7B-Qwen2.
FSR is applied in a fully training-free, plug-and-play manner at inference time, without modifying any model weights.

\textbf{Implementation Details.}
All experiments were implemented using PyTorch 2.1.2 and Python 3.10 with CUDA 12.4.
Regarding hardware configurations, experiments on 7B parameter models were conducted on NVIDIA GeForce RTX 3090 (24GB).
Experiments involving larger architectures(13B) and video models (LLaVA-Video-7B-Qwen2) were performed on NVIDIA GPUs with 48GB memory.
The default hyperparameters for FSR are set as follows: $\alpha=3$, $\beta=1$, $\rho=0.9$, and $\kappa=1$, unless otherwise specified.

\textbf{Evaluation benchmarks.}
We conduct experiments on comprehensive benchmarks spanning image and video tasks.
For image understanding, we cover open-ended QA (VQAv2~\cite{vqav2}), compositional reasoning (GQA~\cite{gqa}, ScienceQA~\cite{sqa}), OCR (TextVQA~\cite{textvqa}), and general capability assessment (POPE~\cite{pope}, MME~\cite{mme}, MMBench~\cite{mmb}, MM-Vet~\cite{mmvet}).
For video understanding, we employ three recent benchmarks: MLVU~\cite{zhou2025mlvubenchmarkingmultitasklong} for multi-task long video analysis, MVBench~\cite{li2024mvbenchcomprehensivemultimodalvideo} for fine-grained temporal perception, and Video-MME~\cite{fu2025videommefirstevercomprehensiveevaluation} for comprehensive multimodal evaluation. To further assess expert-level and world-model-oriented video understanding, we additionally evaluate on MMVU~\cite{zhao2025mmvu} and MMWorld~\cite{he2024mmworld}.
To ensure fair comparison, we standardize the evaluation setup by strictly applying the same prompts, post-processing steps, and metrics across all models.

\subsection{Main Results}
\label{subsec:main_results}
\subsubsection{FSR for Standard Benchmarks}
We first evaluate FSR on LLaVA-1.5-7B, a widely adopted benchmark model for visual token pruning. Table~\ref{tab:llava-1.5} presents the performance of different pruning methods under three token budgets: retaining 192, 128, and 64 visual tokens, corresponding to reduction ratios of 66.7\%, 77.8\%, and 88.9\%, respectively.
When retaining 192 tokens (66.7\% reduction), most pruning methods preserve competitive performance. FSR achieves the highest average score of 99.1\%, outperforming strong baselines such as CDPruner (98.5\%) and VisPruner (98.2\%), incurring negligible performance drop compared to the full token set.
As the token budget tightens to 128 tokens (77.8\% reduction),  FSR maintains a robust average of 98.3\%, with gains of 0.7\% and 1.6\% over CDPruner and VisPruner, respectively.

When the budget is further reduced to 64 tokens (88.9\% reduction), FSR demonstrates superior stability.
In this extreme setting, while attention-based methods suffer severe degradation and joint-strategy methods struggle to balance informativeness, FSR consistently maintains its lead, preserving 96.1\% of the original performance and outperforming CDPruner (95.7\%) and VisPruner (93.5\%).
This robustness is particularly evident in complex reasoning tasks.
Specifically, on complex benchmarks requiring holistic understanding and reasoning, such as MMVet and MMBench-EN, FSR consistently outperforms baselines under high compression (e.g., on MMVet with 64 tokens, 32.6 vs. 29.6 for CDPruner).
This indicates that our strategy effectively balances salient local details with background context, preventing information fragmentation and preserving the semantic completeness for complex tasks.

\subsubsection{FSR for High-Resolution Inputs}
\label{subsec:high_res}

Modern VLMs increasingly adopt high-resolution encoders to capture fine-grained details, leading to a massive increase in visual tokens and substantial spatial redundancy.
To evaluate the scalability of our method, we apply FSR to LLaVA-NeXT-7B.
Following prior work~\cite{cdpruner}, we fix the input resolution to 672$\times$672, resulting in 2,880 visual tokens.
As shown in Table~\ref{tab:llava-next}, when retaining 960 tokens (66.7\% reduction), FSR achieves performance comparable to the full-token upper bound (100.0\% retention), effectively eliminating massive redundancy.
As the reduction ratio increases to retaining 640 tokens (77.8\% reduction), FSR remains the top performer, retaining 99.9\% of the original performance.

Even under the most aggressive setting of retaining 320 tokens (88.9\% reduction), FSR continues to lead with 97.6\% performance retention, consistently surpassing CDPruner (97.3\%) and VisPruner (95.4\%).
This result highlights that FSR is particularly well-suited for high-resolution scenarios.
Unlike low-resolution inputs where details are blurred, high-resolution images provide sharper fine-grained features.
FSR effectively capitalizes on this by accurately capturing these clearer local evidences during the Focus stage, while the Scan and Refine stages ensure the preservation of the global context.
Compared to other approaches, FSR's dynamic allocation proves more effective in leveraging the clarity of high-resolution features to maintain high accuracy even with a limited token budget.

\begin{table*}[t]
  \centering
  \caption{Performance comparison of different pruning methods on Qwen2.5-VL-7B. \textbf{Avg.} represents the average relative performance maintained across all tested benchmarks compared to the unpruned baseline. The best results are highlighted in \textbf{bold}.}
  \label{tab:qwen2.5vl}
  \resizebox{\linewidth}{!}{
    \begin{tabular}{l|cccccccc|c}
    \toprule
    \textbf{Method} & \textbf{GQA}& \textbf{$\text{SQA}^\text{IMG}$} & \textbf{$\text{VQA}^\text{Text}$} & \textbf{POPE} & \textbf{MME} & \textbf{$\text{MMB}^\text{EN}$} & \textbf{$\text{MMB}^\text{CN}$} & \textbf{MMVet} & \textbf{Avg.} \\
    \rowcolor{lightgray!75}\multicolumn{10}{c}{\textit{Upper Bound: All Tokens} ($\mathbf{100\%}$)} \\
    \rowcolor{lightgray!25} Qwen2.5-VL-7B & 60.8 & 88.9 & 77.6 & 86.5 & 2328 & 83.5 & 81.4 & 64.4 & 100.0\%\\
    \rowcolor{lightgray!75}\multicolumn{10}{c}{\textit{Reduction Ratio:} \textcolor{Green}{$\downarrow\mathbf{80\%}$}} \\
    FastV{\small\texttt{(ECCV24)}}& 56.8  & 83.1  & 70.7  & 81.0  & 2102 & 76.8  & 75.4  & 57.4 & 92.0\%\\
    HoloV{\small\texttt{(NIPS2025)}} & 59.5  & 87.8  & 73.8  & 85.1  & 2179 & 81.1  & 78.9  & 55.5 & 95.6\%\\
    \ours & 60.2  & 87.9  & 76.0  & 86.1  & 2258 & 81.5  & 79.1  & 61.7 & \textbf{97.9\%}\\
    
    \rowcolor{lightgray!75}\multicolumn{10}{c}{\textit{Reduction Ratio:} \textcolor{Green}{$\downarrow\mathbf{60\%}$}} \\
    FastV{\small\texttt{(ECCV24)}} &  56.3  & 83.1  & 68.8  & 80.2  & 2063 & 75.7  & 73.5  & 51.4 & 89.8\%\\
    HoloV{\small\texttt{(NIPS2025)}} & 59.0  & 87.2  & 71.9  & 84.4  & 2177 & 79.7  & 77.8  & 52.1 & 94.2\%\\
    \ours & 59.9  & 87.5  & 75.1  & 85.2  & 2227 & 80.3  & 78.5  & 57.5 & \textbf{96.4\%}\\
    
    \rowcolor{lightgray!75}\multicolumn{10}{c}{\textit{Reduction Ratio:} \textcolor{Green}{$\downarrow\mathbf{80\%}$}} \\
    FastV{\small\texttt{(ECCV24)}} & 54.2  & 82.2  & 61.0  & 77.5  & 1915 & 72.5  & 70.0  & 44.7 & 84.6\%\\
    HoloV{\small\texttt{(NIPS2025)}} &57.1  & 86.0  & 64.5  & 81.3  & 2008 & 76.3  & 73.4  & 45.3 & 88.6\%\\
    \ours & 58.3  & 86.7  & 70.3  & 83.2  & 2089 & 78.7  & 74.9  & 49.8 & \textbf{91.9\%}\\
    
    \rowcolor{lightgray!75}\multicolumn{10}{c}{\textit{Reduction Ratio:} \textcolor{Green}{$\downarrow\mathbf{90\%}$}} \\
    FastV{\small\texttt{(ECCV24)}} & 50.8  & 80.0  & 53.0  & 72.2  & 1794.7 & 68.2  & 65.1  & 37.1 & 78.3\%\\
    HoloV{\small\texttt{(NIPS2025)}}& 53.6  & 84.4  & 55.7  & 76.4  & 1831 & 72.3  & 68.9  & 38.9 & 82.1\%\\
    \ours & 54.1  & 84.5  & 61.0  & 77.3  & 1907 & 71.7  & 68.3  & 41.4 & \textbf{84.0\%}\\
    \bottomrule
    \end{tabular}
  }
\end{table*}

\begin{table*}[t]
  \centering
  \caption{Performance comparison of different pruning methods on LLaVA-Video-7B-qwen2 with 32 frames per video. \textbf{Avg.} represents the average percentage of performance maintained. ``w/o'' and ``w/'' indicate without and with subtitles.}
  \label{tab:llava-video}
    \resizebox{\linewidth}{!}{
    \begin{tabular}{l|ccccccc|c}
    \toprule
    \textbf{Method} & \textbf{MMVU} & \textbf{MMWorld} &\textbf{MLVU} & \textbf{MVBench} & \multicolumn{3}{c|}{\textbf{Video-MME}} & \\
    \textbf{Metric} & \textbf{val}& \textbf{test} & \textbf{test} & \textbf{test} & \textbf{all+w/o} & \textbf{all+w/} & \textbf{long} & \textbf{Avg.} \\
    \rowcolor{lightgray!75}\multicolumn{9}{c}{\textit{Upper Bound: All Tokens} ($\mathbf{100\%}$)} \\
    \rowcolor{lightgray!25} \small{LLaVA-Video-7B-qwen2}& 44.0 & 30.0 & 50.1 & 60.8 & 62.6 & 62.4 & 51.8 & 100\% \\
    \rowcolor{lightgray!75}\multicolumn{9}{c}{\textit{Reduction Ratio:} \textcolor{Green}{$\downarrow\mathbf{50\%}$}}   \\
    HoloV{\small\texttt{(NIPS2025)}}& 44.2 & 31.5 & 49.1 & 59.4 & 61.7 & 61.6 & 51.3 & 99.2\%\\
    \ours & 46.0 & 31.1 & 50.2 & 59.7 & 61.9 & 62.0 & 51.6 & \textbf{100.3\%}\\
    \rowcolor{lightgray!75}\multicolumn{9}{c}{\textit{Reduction Ratio:} \textcolor{Green}{$\downarrow\mathbf{60\%}$}}   \\
    HoloV{\small\texttt{(NIPS2025)}}& 43.4 & 30.8 & 49.1 & 59.3 & 61.4 & 61.0 & 51.3 & 98.5\%\\
    \ours &44.6 & 31.1& 50.0 & 59.4 & 61.6 & 61.5 & 52.2 & \textbf{99.6\%} \\
    \rowcolor{lightgray!75}\multicolumn{9}{c}{\textit{Reduction Ratio:} \textcolor{Green}{$\downarrow\mathbf{70\%}$}}\\
    HoloV{\small\texttt{(NIPS2025)}}& 43.7 & 31.0& 48.5 & 59.0 & 60.6 & 61.2 & 51.2 & 98.2\%\\
    \ours & 44.6 & 31.6 & 47.6 & 59.2 & 61.3 & 61.5 & 52.0 & \textbf{98.9\%}\\
    \rowcolor{lightgray!75}\multicolumn{9}{c}{\textit{Reduction Ratio:} \textcolor{Green}{$\downarrow\mathbf{80\%}$}}  \\
    HoloV{\small\texttt{(NIPS2025)}}&44.0 & 32.9 & 46.5 & 58.3 & 60.4 & 60.8 & 51.6 & 98.0\%\\
    \ours & 43.4 & 33.3 & 46.5 & 58.5 & 60.2 & 60.9 & 52.3 & \textbf{98.2\%}\\
    \bottomrule
    \end{tabular}
}
\end{table*}

\begin{table*}[t]
  \centering
  \caption{Performance comparison of different pruning methods on LLaVA-1.5-13B. \textbf{Avg.} represents the average relative performance maintained across all tested benchmarks compared to the unpruned baseline. The best results are highlighted in \textbf{bold}.}
  \label{tab:llava-1.5-13b}
  \resizebox{\linewidth}{!}{
    \begin{tabular}{l|ccccccccc|c}
    \toprule
    \textbf{Method} & \textbf{$\text{VQA}^\text{V2}$} & \textbf{GQA} & \textbf{$\text{SQA}^\text{IMG}$} & \textbf{$\text{VQA}^\text{Text}$} & \textbf{POPE} & \textbf{MME} & \textbf{$\text{MMB}^\text{EN}$} & \textbf{$\text{MMB}^\text{CN}$} & \textbf{MMVet} & \textbf{Avg.} \\
    \rowcolor{lightgray!75}\multicolumn{11}{c}{\textit{Upper Bound, All 576 Tokens} ($\mathbf{100\%}$)} \\
    \rowcolor{lightgray!25} LLaVA-1.5-13B & 80.0 & 63.3 & 72.8 & 61.2 & 86.1 & 1828 & 68.5 & 63.5 & 36.7 & 100\% \\
    \rowcolor{lightgray!75}\multicolumn{11}{c}{\textit{Retain 192 Tokens} \textcolor{Green}{($\downarrow\mathbf{66.7\%}$)}} \\
     HoloV{\small\texttt{(NIPS2025)}} & - & 58.5 & 72.3 & 58.0 & 84.2 & 1754 & 65.8 & 60.2 & 35.5 & 96.1\% \\
     VisPruner{\small\texttt{(ICCV2025)}} & 78.1 & 59.5 & 73.9 & 59.7 & 86.0 & 1750 & 67.2 & 62.5 & 37.0 & 98.1\% \\
     CDPruner{\small\texttt{(NIPS2025)}} & 78.4 & 60.4 & 72.4 & 58.7 & 86.6 & 1776 & 67.2 & 62.1 & 35.7  & 97.9\% \\
     \ours  & 78.6 & 60.2 & 73.3 & 59.5 & 86.4 & 1805 & 67.3 & 63.0 & 37.4 & \textbf{98.8\%} \\
    \rowcolor{lightgray!75}\multicolumn{11}{c}{\textit{Retain 128 Tokens} \textcolor{Green}{($\downarrow\mathbf{77.8\%}$)}} \\
     FastV{\small\texttt{(ECCV24)}} & 75.3 & 58.3  & 74.2 & 58.6 & 75.5 & 1722 & 66.1 & 62.3 & 32.8 &  94.5\% \\
     VisionZip{\small\texttt{(CVPR25)}} & 76.8 & 57.9 & 73.8 & 58.9 & 82.7 & 1710 & 67.4 & 62.5 & 36.0 &  96.5\% \\
     DivPrune{\small\texttt{(CVPR25)}} & 77.1 & 59.2  & 72.8 & 58.0 & 86.8 & 1720 & 66.3 & 60.7 & 34.4 &  96.4\% \\
     HoloV{\small\texttt{(NIPS2025)}} & - & 57.5 & 73.6 & 58.1 & 81.9 & 1731 & 66.5 & 62.0 & 35.4 & 96.0\% \\
     VisPruner{\small\texttt{(ICCV2025)}} & 76.9 & 58.4 & 73.9 & 59.2 & 83.8 & 1736 & 67.2 & 62.2 & 36.9 & 97.1\% \\
     CDPruner{\small\texttt{(NIPS2025)}} & 77.7 & 59.7 & 72.5 & 58.4 & 87.3 & 1778 & 67.5 & 61.4 & 37.3 & 97.9\% \\
     \ours  & 78.0 & 59.6 & 73.8 & 58.8 & 86.3 & 1768 & 68.2 & 61.6 & 38.8 & \textbf{98.4\%} \\
    \rowcolor{lightgray!75}\multicolumn{11}{c}{\textit{Retain 64 Tokens} \textcolor{Green}{($\downarrow\mathbf{88.9\%}$)}} \\
     FastV{\small\texttt{(ECCV24)}} & 65.3 & 51.9  & 73.1 & 53.4 & 56.9 & 1470 & 59.2 & 55.1 & 26.9  &  82.6\% \\
     VisionZip{\small\texttt{(CVPR25)}} & 73.7 & 56.2  & 74.2 & 57.4 & 75.7 & 1628 & 64.9 & 61.3 & 33.4 & 92.7\% \\
     DivPrune{\small\texttt{(CVPR25)}} & 75.2 & 57.9 & 71.7 & 57.4 & 84.5 & 1713 & 64.1 & 59.8 & 29.3 & 93.9\% \\
     HoloV{\small\texttt{(NIPS2025)}} & - & 56.0 & 74.2 & 57.1 & 75.6 & 1683 & 64.4 & 60.3 & 33.9 & 93.0\% \\
     VisPruner{\small\texttt{(ICCV2025)}}& 73.9 & 56.0 & 74.0 & 57.9 & 79.2 & 1694  & 65.0 & 59.9 & 33.1 & 93.6\% \\
     CDPruner{\small\texttt{(NIPS2025)}} & 76.7 & 59.4 & 72.4 & 57.6 & 87.2 & 1744 & 65.5 & 58.9 & 35.8 & 96.3\% \\
     \ours & 76.8 & 58.6 & 73.0 & 58.1 & 85.0 & 1750 & 66.3 & 60.7 & 36.7 & \textbf{96.7\%} \\
    \bottomrule
    \end{tabular}
  }
\end{table*}

\begin{table*}[t]
  \centering
  \caption{Performance comparison of different pruning methods on LLaVA-NeXT-13B. \textbf{Avg.} represents the average relative performance maintained across all tested benchmarks compared to the unpruned baseline. The best results are highlighted in \textbf{bold}.}
  \label{tab:llava-next-13b}
  \resizebox{\linewidth}{!}{
    \begin{tabular}{l|ccccccccc|c}
    \toprule
    \textbf{Method} & \textbf{$\text{VQA}^\text{V2}$} & \textbf{GQA} & \textbf{$\text{SQA}^\text{IMG}$} & \textbf{$\text{VQA}^\text{Text}$} & \textbf{POPE} & \textbf{MME} & \textbf{$\text{MMB}^\text{EN}$} & \textbf{$\text{MMB}^\text{CN}$} & \textbf{MMVet} &  \textbf{Avg.} \\
    \rowcolor{lightgray!75}\multicolumn{11}{c}{\textit{Upper Bound, All 2880 Tokens} ($\mathbf{100\%}$)} \\
    \rowcolor{lightgray!25} LLaVA-NeXT-13B & 82.3 & 64.3 & 73.2 & 63.2 & 85.3 & 1837 & 68.6 & 61.2 & 36.6 & 100.0\% \\
    \rowcolor{lightgray!75}\multicolumn{11}{c}{\textit{Retain 960 Tokens} \textcolor{Green}{($\downarrow\mathbf{66.7\%}$)}} \\
     HoloV{\small\texttt{(NIPS2025)}}& - & 63.1 & 69.5 & 60.6 & 85.8 & 1840 & 64.4 & 56.4 & 42.3 & 98.1\% \\
     VisPruner{\small\texttt{(ICCV2025)}}& 80.8 & 63.7 & 71.9 & 62.5 & 86.0 & 1902 & 69.0 & 63.1 & 45.0 & 101.7\% \\
     CDPruner{\small\texttt{(NIPS2025)}}& 81.6 & 64.3 & 72.1 & 61.3 & 87.1 & 1880 & 69.2 & 62.5 & 41.7 & 101.2\% \\
     \ours & 81.3 & 64.4 & 72.2 & 62.5 & 86.9 & 1885 & 70.4 & 63.3 & 44.7 & \textbf{102.1\%} \\

    \rowcolor{lightgray!75}\multicolumn{11}{c}{\textit{Retain 640 Tokens} \textcolor{Green}{($\downarrow\mathbf{77.8\%}$)}} \\
     FastV{\small\texttt{(ECCV24)}} & 79.4 & 60.9  & 71.7 & 60.7 & 80.2 & 1804 & 65.5 & 59.9 & 43.8 & 97.7\% \\
     VisionZip{\small\texttt{(CVPR25)}} & 79.7 & 62.9 & 70.8 & 62.1 & 85.8 & 1844 & 68.1 & 62.6 & 46.8 & 100.7\% \\
     DivPrune{\small\texttt{(CVPR25)}} & 80.4 & 63.5  & 72.2 & 59.2 & 86.5 & 1816 & 67.5 & 62.9 & 39.0 & 99.3\% \\
     HoloV{\small\texttt{(NIPS2025)}}& - & 62.8 & 71.7 & 60.0 & 85.9 & 1830 & 67.0 & 60.6 & 41.3 & 99.4\% \\
     VisPruner{\small\texttt{(ICCV2025)}}& 79.7 & 62.9 & 71.1 & 62.0 & 84.6 & 1876 & 67.7 & 62.6 & 46.1 & 100.6\% \\
     CDPruner{\small\texttt{(NIPS2025)}}& 81.0 & 63.9 & 71.9 & 60.9 & 87.6 & 1871 & 68.9 & 62.5 & 41.4 & 100.8\% \\
     \ours & 80.6 & 63.8 & 71.6 & 61.9 & 87.2 & 1908 & 69.2 & 63.3 & 44.1 & \textbf{101.7\%} \\
    
    \rowcolor{lightgray!75}\multicolumn{11}{c}{\textit{Retain 320 Tokens} \textcolor{Green}{($\downarrow\mathbf{88.9\%}$)}} \\
     FastV{\small\texttt{(ECCV24)}} & 69.8 & 54.6 & 70.5 & 55.4 & 63.6 & 1522 & 59.8 & 54.4 & 30.2 & 85.3\% \\
     VisionZip{\small\texttt{(CVPR25)}} & 76.8 & 60.7 & 70.2 & 60.7 & 82.3 & 1770 & 66.5 & 62.3 & 41.1 & 97.2\% \\
     DivPrune{\small\texttt{(CVPR25)}} & 78.1 & 61.8 & 72.3 & 57.6 & 85.2 & 1753 & 65.9 & 61.9 & 39.2 & 97.3\% \\
     HoloV{\small\texttt{(NIPS2025)}}& - & 60.9 & 70.6 & 59.5 & 83.4 & 1789 & 67.9 & 62.5 & 40.7 & 98.3\% \\
     VisPruner{\small\texttt{(ICCV2025)}}& 76.7 & 60.4 & 70.5 & 60.3 & 81.2 & 1831 & 66.5 & 62.5 & 39.7 & 97.3\% \\
     CDPruner{\small\texttt{(NIPS2025)}}& 79.5 & 63.0 & 71.1 & 59.0 & 87.6 & 1789 & 66.8 & 61.9 & 42.1 & 99.0\% \\
     \ours & 78.8 & 62.7 & 70.3 & 60.3 & 86.8 & 1882  & 67.9 & 63.1 & 42.3 & \textbf{100.0\%} \\

    \bottomrule
    \end{tabular}
  }
\end{table*}

\begin{table*}[t]
    \centering
    \caption{Comparison of efficiency and performance metrics on LLaVA-1.5-7B. We evaluate computational cost, inference latency, and memory footprint when retaining 64 visual tokens. \textbf{Score} denotes the accuracy performance on the MMBench-EN benchmark.}
    \label{tab:efficiency}

    \setlength{\tabcolsep}{2.5pt} 
    \resizebox{\linewidth}{!}{
        \begin{tabular}{lcccccccc} 
            \toprule
            \multirow{2}{*}{\textbf{Method}} & \multirow{2}{*}{\textbf{Token}} & \textbf{FLOPS} & \textbf{MACs}& \textbf{Prefill Time} & \textbf{Decode Time} & \textbf{KV Cache} & \textbf{GPU Memory} & \multirow{2}{*}{\textbf{Score}} \\
             & & (T) & (T) &(ms/token) & (ms/token) & (MB) & (GB) & \\
            \midrule
            LLaVA-1.5-7B & 576 & 9.042 & 4.521 & 3.056 & 23.952 & 288.0 & 16.9 & 64.6 \\
            \midrule
            FastV        & 64  & 3.610 & 1.800 & 0.912 & 23.123 & 32.1 & 15.4 & 50.1 \\
            VisPruner    & 64  & 2.277 & \textbf{1.138} & 0.791 & 22.573 & 32.0 & 13.2 & 59.4 \\
            CDPruner     & 64  & 2.293 & 1.146 &\textbf{0.775} & 22.563 & 32.0 & 13.2 & 60.8 \\
            Ours & 64 & \textbf{2.291} & 1.145 & 0.788 & \textbf{22.317} & \textbf{32.0} & \textbf{13.2} & \textbf{61.9} \\
            \bottomrule
        \end{tabular}
    }
\end{table*}

\begin{figure*}[!t]
\centering
\includegraphics[width=\textwidth]{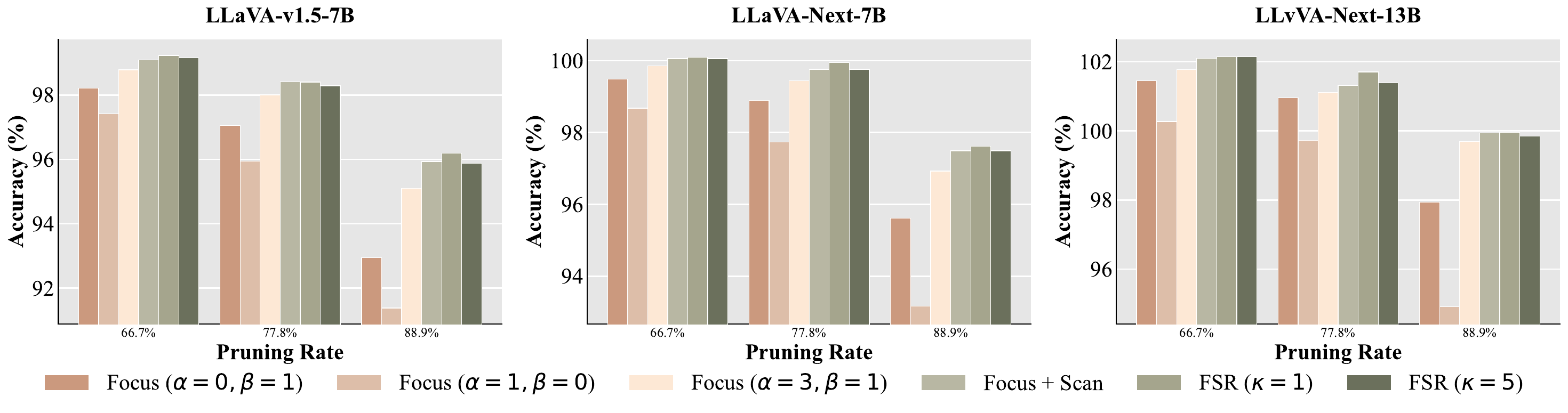}
\caption{Ablation study on LLaVA-1.5-7B, LLaVA-NeXT-7B, and LLaVA-NeXT-13B across varying pruning ratios, validating the impact of dual-pathway hyperparameters ($\alpha, \beta$), focus-conditioned scanning, and aggregation refinement ratio ($\kappa$).}
\label{fig:ablation}
\end{figure*}

\subsubsection{FSR for Advanced Architectures}
\label{subsec:advanced}

To further evaluate the generality of FSR beyond LLaVA-style architectures, we conduct experiments on Qwen2.5-VL-7B, a more advanced VLM that supports dynamic image resolution and native token merging.
These built-in efficiency designs inherently reduce token redundancy, making training-free token pruning more challenging in practice.
Despite this stronger baseline, FSR still achieves the best accuracy--efficiency trade-off.
To ensure a fair and architecture-compatible evaluation, we apply a minimal adaptation of FSR to Qwen2.5-VL-7B:
the Focus-stage scores are derived by aggregating the self-attention map of visual tokens, and the instruction relevance term is omitted due to the absence of text encoder.

Table~\ref{tab:qwen2.5vl} reports the results under different token reduction ratios, ranging from moderate (50\%, 60\%) to aggressive (80\%, 90\%) pruning.
Across all reduction ratios, FSR consistently outperforms representative baselines, including FastV and HoloV.
Under moderate compression (50\% and 60\%), FSR preserves nearly all of the original performance, achieving average scores of 97.9\% and 96.4\%, respectively, while maintaining clear margins over competing methods.
As the compression ratio increases, the advantage of FSR becomes more pronounced.
With 80\% of visual tokens removed, FSR retains 91.9\% performance, surpassing HoloV by 3.3\%.
At the extreme setting of 90\% token reduction, FSR still achieves 84.0\% of the original performance, compared to 82.1\% for HoloV and 78.3\% for FastV.

The benefits of FSR are particularly evident on benchmarks that require integrated multimodal reasoning and robust global understanding.
For example, on MMVet and MME, FSR consistently maintains superior performance even under aggressive compression, demonstrating its exceptional robustness in preserving critical information for complex reasoning tasks.

\subsubsection{FSR for Video Understanding}
\label{subsec:video}

We further assess the generalization of FSR to the video domain on LLaVA-Video-7B-Qwen2, utilizing 32 frames per video to capture temporal dynamics.
As presented in Table~\ref{tab:llava-video}, FSR consistently outperforms the state-of-the-art method HoloV across varying pruning ratios ranging from 50\% to 80\%.
Notably, at 60\% pruning ratio, FSR retains 99.6\% of the original performance, significantly surpassing HoloV (98.5\%) and effectively serving as a highly efficient substitute for the full token set.
Even under aggressive compression where 80\% of tokens are removed, FSR demonstrates superior robustness, maintaining an average score of 98.2\% compared to 98.0\% for HoloV.
This indicates that FSR's strategy of balancing local evidence and global context effectively extends to the temporal dimension, enabling robust preservation of critical spatiotemporal cues in challenging benchmarks.

\subsubsection{FSR for Large-Scale Models}
\label{subsec:large}
We further evaluate the effectiveness of FSR on larger scale VLMs, including LLaVA-1.5-13B and the more advanced LLaVA-NeXT-13B.
The results are summarized in Tables~\ref{tab:llava-1.5-13b} and \ref{tab:llava-next-13b}, respectively, under multiple token budgets ranging from moderate to aggressive pruning.

On LLaVA-1.5-13B, FSR consistently achieves the best accuracy--efficiency trade-off across all pruning ratios.
Even with 88.9\% of visual tokens removed, FSR retains 96.7\% of the original performance, clearly outperforming representative baselines such as VisPruner and CDPruner.
More notably, on LLaVA-NeXT-13B, FSR exhibits an interesting behavior.
When retaining only 640 visual tokens (77.8\% reduction), FSR slightly outperforms the unpruned baseline, achieving an average score of 101.7\%.
This result suggests that the original dense visual token set contains substantial redundancy, which may introduce noise and interfere with multimodal reasoning.
By selectively preserving informative local evidence while maintaining sufficient global context, FSR effectively filters out distracting tokens, leading to more focused and accurate reasoning.

\subsection{Efficiency Analysis}
\label{subsec:efficiency}

We evaluate the efficiency of FSR in terms of computational cost, inference latency, and memory footprint on a single NVIDIA RTX 3090 GPU.
As shown in Table~\ref{tab:efficiency}, retaining only 64 tokens, FSR yields substantial resource savings compared to the LLaVA-1.5-7B baseline: FLOPs are reduced by approximately 75\%, and KV cache memory is compressed by nearly $\mathbf{9\times}$.
These reductions translate into significant runtime benefits, achieving a $\mathbf{3.9\times}$ speedup in the prefill stage.

Crucially, FSR achieves the most superior accuracy--efficiency trade-off among all compared methods.
FSR maintains the lowest decode latency (22.317 ms) and matches the prefill speed of state-of-the-art pruners like CDPruner, confirming that our pipeline introduces negligible system overhead.
While purely efficiency-oriented methods like FastV suffer severe accuracy drops, FSR delivers the highest score in MMBench-EN, validating its suitability for practical, high-performance deployment.

\subsection{Ablation Study}
\label{subsec:ablation}

We conduct ablation studies on LLaVA-v1.5-7B, LLaVA-NeXT-7B, and LLaVA-NeXT-13B to examine the contribution of each component in FSR across varying pruning ratios.
The results are summarized in Figure~\ref{fig:ablation}.
Starting from single-cue baselines, we progressively validate the efficacy of the proposed Focus--Scan--Refine pipeline.

\textbf{Impact of hyperparameters $\alpha$ and $\beta$.}
We first investigate the trade-off between instruction relevance ($\hat{r}$) and visual saliency ($\hat{s}$) by varying the exponents in Eq.~\ref{eq:fused_score} ($\phi_i=\hat{r}_i^{\alpha}\hat{s}_i^{\beta}$).
As shown in Figure~\ref{fig:ablation}, relying solely on visual saliency ($\alpha=0, \beta=1$) or instruction relevance ($\alpha=1, \beta=0$) leads to noticeable performance degradation, especially under aggressive reduction (88.9\%).
For instance, instruction relevance alone often fails to capture background context, while visual saliency may miss task-specific targets.
In contrast, the dual-pathway strategy ($\alpha=3, \beta=1$) consistently achieves the highest accuracy across all models.
This demonstrates that visual saliency and semantic relevance provide complementary signals—one capturing intrinsic visual prominence and the other ensuring instruction-level alignment.

\textbf{Effectiveness of focus-conditioned scan.}
Building upon the dual-pathway selection, introducing the second-stage Scan mechanism boosts performance.
Compared to using focused tokens alone, this stage effectively supplements complementary global context conditioned on the local evidence.
This addition proves crucial for multi-object understanding and reasoning-heavy queries where local cues are insufficient.
Notably, the performance gains are most pronounced under aggressive compression, where the information captured by the Focus stage becomes limited and the Scan stage plays a critical role in supplementing sufficient global context.

\textbf{Impact of aggregation refinement.}
The Refine stage provides a further performance boost, which becomes increasingly valuable under extreme reduction ratios.
By aggregating discarded but relevant tokens into the scan anchors, FSR recovers missing details without expanding the token budget.
However, we observe that the gain saturates when the merge ratio is excessive ($\kappa=5$), as merging too many tokens tends to blur the aggregated representation.
A moderate refine ratio ($\kappa=1$) achieves the optimal trade-off, delivering consistent gains by enriching context without over-smoothing features.
Interestingly, we note that this benefit is less pronounced on larger models like LLaVA-NeXT-13B, suggesting that stronger LLM backbones possess higher tolerance for minor information loss in peripheral regions.


\section{Conclusion}

In this paper, we propose FSR, a training-free visual token pruning framework inspired by human visual perception, which addresses the fundamental challenge of allocating a limited token budget in VLMs.
FSR explicitly models the progressive coordination between local evidence and global context through a three-stage process: focusing on task-critical regions, scanning for complementary contextual cues, and refining sparse representations via aggregation.
By jointly considering visual saliency, conditional global coverage, and redundancy-aware refinement, FSR preserves both query-relevant evidence and holistic scene information under strict token constraints.

Extensive experiments across diverse model architectures, input resolutions, and image--video benchmarks demonstrate that FSR consistently achieves a superior accuracy--efficiency trade-off compared to prior methods.
These results highlight the effectiveness of human-inspired local--global coordination as a general paradigm for efficient multimodal inference, and position FSR as a practical solution for deploying large-scale VLMs under real-world resource constraints.

\section*{Statements and Declarations}

\textbf{Competing Interests.}
The authors declare that they have no competing interests.

\noindent\textbf{Data Availability.} All data analyzed during this study are included in this published article. The original publicly available datasets used for evaluation are cited within the manuscript.


\bibliography{sn-bibliography}

@misc{openai2024gpt4technicalreport,
      title={GPT-4 Technical Report}, 
      author={OpenAI and Josh Achiam and Steven Adler and Sandhini Agarwal and Lama Ahmad and Ilge Akkaya and Florencia Leoni Aleman and Diogo Almeida and Janko Altenschmidt and Sam Altman and Shyamal Anadkat and et al},
      year={2024},
      eprint={2303.08774},
      archivePrefix={arXiv},
      primaryClass={cs.CL},
      url={https://arxiv.org/abs/2303.08774}, 
}

@article{touvron2023llama,
  title={Llama: Open and efficient foundation language models},
  author={Touvron, Hugo and Lavril, Thibaut and Izacard, Gautier and Martinet, Xavier and Lachaux, Marie-Anne and Lacroix, Timoth{\'e}e and Rozi{\`e}re, Baptiste and Goyal, Naman and Hambro, Eric and Azhar, Faisal and others},
  journal={arXiv preprint arXiv:2302.13971},
  year={2023}
}

@misc{jiang2023mistral7b,
      title={Mistral 7B}, 
      author={Albert Q. Jiang and Alexandre Sablayrolles and Arthur Mensch and Chris Bamford and Devendra Singh Chaplot and Diego de las Casas and Florian Bressand and Gianna Lengyel and Guillaume Lample and Lucile Saulnier and Lélio Renard Lavaud and Marie-Anne Lachaux and Pierre Stock and Teven Le Scao and Thibaut Lavril and Thomas Wang and Timothée Lacroix and William El Sayed},
      year={2023},
      eprint={2310.06825},
      archivePrefix={arXiv},
      primaryClass={cs.CL},
      url={https://arxiv.org/abs/2310.06825}, 
}

@misc{qwen2025qwen25technicalreport,
      title={Qwen2.5 Technical Report}, 
      author={Qwen and : and An Yang and Baosong Yang and Beichen Zhang and Binyuan Hui and Bo Zheng and Bowen Yu and Chengyuan Li and Dayiheng Liu and Fei Huang and Haoran Wei and Huan Lin and Jian Yang and Jianhong Tu and Jianwei Zhang and Jianxin Yang and Jiaxi Yang and Jingren Zhou and Junyang Lin and Kai Dang and Keming Lu and Keqin Bao and Kexin Yang and Le Yu and Mei Li and Mingfeng Xue and Pei Zhang and Qin Zhu and Rui Men and Runji Lin and Tianhao Li and Tianyi Tang and Tingyu Xia and Xingzhang Ren and Xuancheng Ren and Yang Fan and Yang Su and Yichang Zhang and Yu Wan and Yuqiong Liu and Zeyu Cui and Zhenru Zhang and Zihan Qiu},
      year={2025},
      eprint={2412.15115},
      archivePrefix={arXiv},
      primaryClass={cs.CL},
      url={https://arxiv.org/abs/2412.15115}, 
}

@misc{clip,
      title={Learning Transferable Visual Models From Natural Language Supervision}, 
      author={Alec Radford and Jong Wook Kim and Chris Hallacy and Aditya Ramesh and Gabriel Goh and Sandhini Agarwal and Girish Sastry and Amanda Askell and Pamela Mishkin and Jack Clark and Gretchen Krueger and Ilya Sutskever},
      year={2021},
      eprint={2103.00020},
      archivePrefix={arXiv},
      primaryClass={cs.CV},
      url={https://arxiv.org/abs/2103.00020}, 
}

@misc{alayrac2022flamingo,
      title={Flamingo: a Visual Language Model for Few-Shot Learning}, 
      author={Jean-Baptiste Alayrac and Jeff Donahue and Pauline Luc and Antoine Miech and Iain Barr and Yana Hasson and Karel Lenc and Arthur Mensch and Katie Millican and Malcolm Reynolds and Roman Ring and Eliza Rutherford and Serkan Cabi and Tengda Han and Zhitao Gong and Sina Samangooei and Marianne Monteiro and Jacob Menick and Sebastian Borgeaud and Andrew Brock and Aida Nematzadeh and Sahand Sharifzadeh and Mikolaj Binkowski and Ricardo Barreira and Oriol Vinyals and Andrew Zisserman and Karen Simonyan},
      year={2022},
      eprint={2204.14198},
      archivePrefix={arXiv},
      primaryClass={cs.CV},
      url={https://arxiv.org/abs/2204.14198}, 
}

@misc{li2023blip2,
      title={BLIP-2: Bootstrapping Language-Image Pre-training with Frozen Image Encoders and Large Language Models}, 
      author={Junnan Li and Dongxu Li and Silvio Savarese and Steven Hoi},
      year={2023},
      eprint={2301.12597},
      archivePrefix={arXiv},
      primaryClass={cs.CV},
      url={https://arxiv.org/abs/2301.12597}, 
}

@misc{dai2023instructblip,
      title={InstructBLIP: Towards General-purpose Vision-Language Models with Instruction Tuning}, 
      author={Wenliang Dai and Junnan Li and Dongxu Li and Anthony Meng Huat Tiong and Junqi Zhao and Weisheng Wang and Boyang Li and Pascale Fung and Steven Hoi},
      year={2023},
      eprint={2305.06500},
      archivePrefix={arXiv},
      primaryClass={cs.CV},
      url={https://arxiv.org/abs/2305.06500}, 
}

@misc{zhu2023minigpt4,
      title={MiniGPT-4: Enhancing Vision-Language Understanding with Advanced Large Language Models}, 
      author={Deyao Zhu and Jun Chen and Xiaoqian Shen and Xiang Li and Mohamed Elhoseiny},
      year={2023},
      eprint={2304.10592},
      archivePrefix={arXiv},
      primaryClass={cs.CV},
      url={https://arxiv.org/abs/2304.10592}, 
}

@misc{liu2023llava,
      title={Visual Instruction Tuning}, 
      author={Haotian Liu and Chunyuan Li and Qingyang Wu and Yong Jae Lee},
      year={2023},
      eprint={2304.08485},
      archivePrefix={arXiv},
      primaryClass={cs.CV},
      url={https://arxiv.org/abs/2304.08485}, 
}

@misc{bai2023qwenvl,
      title={Qwen-VL: A Versatile Vision-Language Model for Understanding, Localization, Text Reading, and Beyond}, 
      author={Jinze Bai and Shuai Bai and Shusheng Yang and Shijie Wang and Sinan Tan and Peng Wang and Junyang Lin and Chang Zhou and Jingren Zhou},
      year={2023},
      eprint={2308.12966},
      archivePrefix={arXiv},
      primaryClass={cs.CV},
      url={https://arxiv.org/abs/2308.12966}, 
}

@misc{chen2024internvl,
      title={InternVL: Scaling up Vision Foundation Models and Aligning for Generic Visual-Linguistic Tasks}, 
      author={Zhe Chen and Jiannan Wu and Wenhai Wang and Weijie Su and Guo Chen and Sen Xing and Muyan Zhong and Qinglong Zhang and Xizhou Zhu and Lewei Lu and Bin Li and Ping Luo and Tong Lu and Yu Qiao and Jifeng Dai},
      year={2024},
      eprint={2312.14238},
      archivePrefix={arXiv},
      primaryClass={cs.CV},
      url={https://arxiv.org/abs/2312.14238}, 
}

@misc{gpt4v,
  author = {{OpenAI}},
  title = {GPT-4V(ision) System Card},
  howpublished = {OpenAI Blog},
  year = {2023},
  url = {https://openai.com/research/gpt-4v-system-card},
  note = {Accessed: Jan. 15, 2024}
}

@misc{gemini,
      title={Gemini: A Family of Highly Capable Multimodal Models}, 
      author={Gemini Team and Rohan Anil and Sebastian Borgeaud and Jean-Baptiste Alayrac and Jiahui Yu and Radu Soricut and Johan Schalkwyk and Andrew M. Dai and Anja Hauth and Katie Millican and David Silver and Melvin Johnson and Ioannis Antonoglou and Julian Schrittwieser and Amelia Glaese and Jilin Chen and Emily Pitler and Timothy Lillicrap and et al},
      year={2025},
      eprint={2312.11805},
      archivePrefix={arXiv},
      primaryClass={cs.CL},
      url={https://arxiv.org/abs/2312.11805}, 
}

@misc{li2024llavanext,
      title={LLaVA-NeXT-Interleave: Tackling Multi-image, Video, and 3D in Large Multimodal Models}, 
      author={Feng Li and Renrui Zhang and Hao Zhang and Yuanhan Zhang and Bo Li and Wei Li and Zejun Ma and Chunyuan Li},
      year={2024},
      eprint={2407.07895},
      archivePrefix={arXiv},
      primaryClass={cs.CV},
      url={https://arxiv.org/abs/2407.07895}, 
}

@article{vaswani2017attention,
  title={Attention is all you need},
  author={Vaswani, Ashish and Shazeer, Noam and Parmar, Niki and Uszkoreit, Jakob and Jones, Llion and Gomez, Aidan N and Kaiser, {\L}ukasz and Polosukhin, Illia},
  journal={Advances in neural information processing systems},
  volume={30},
  year={2017}
}

@article{team2024gemma,
  title={Gemma: Open models based on gemini research and technology},
  author={Team, Gemma and Mesnard, Thomas and Hardin, Cassidy and Dadashi, Robert and Bhupatiraju, Surya and Pathak, Shreya and Sifre, Laurent and Rivi{\`e}re, Morgane and Kale, Mihir Sanjay and Love, Juliette and others},
  journal={arXiv preprint arXiv:2403.08295},
  year={2024}
}

@article{hu2024minicpm,
  title={Minicpm: Unveiling the potential of small language models with scalable training strategies},
  author={Hu, Shengding and Tu, Yuge and Han, Xu and He, Chaoqun and Cui, Ganqu and Long, Xiang and Zheng, Zhi and Fang, Yewei and Huang, Yuxiang and Zhao, Weilin and others},
  journal={arXiv preprint arXiv:2404.06395},
  year={2024}
}

@inproceedings{topv,
  title={Topv: Compatible token pruning with inference time optimization for fast and low-memory multimodal vision language model},
  author={Yang, Cheng and Sui, Yang and Xiao, Jinqi and Huang, Lingyi and Gong, Yu and Li, Chendi and Yan, Jinghua and Bai, Yu and Sadayappan, Ponnuswamy and Hu, Xia and others},
  booktitle={Proceedings of the Computer Vision and Pattern Recognition Conference},
  pages={19803--19813},
  year={2025}
}

@inproceedings{fastv,
  title={An image is worth 1/2 tokens after layer 2: Plug-and-play inference acceleration for large vision-language models},
  author={Chen, Liang and Zhao, Haozhe and Liu, Tianyu and Bai, Shuai and Lin, Junyang and Zhou, Chang and Chang, Baobao},
  booktitle={European Conference on Computer Vision},
  pages={19--35},
  year={2024},
  organization={Springer}
}

@misc{prumerge,
      title={LLaVA-PruMerge: Adaptive Token Reduction for Efficient Large Multimodal Models}, 
      author={Yuzhang Shang and Mu Cai and Bingxin Xu and Yong Jae Lee and Yan Yan},
      year={2024},
      eprint={2403.15388},
      archivePrefix={arXiv},
      primaryClass={cs.CV},
      url={https://arxiv.org/abs/2403.15388}, 
}

@inproceedings{fitprune,
  title={Fit and prune: Fast and training-free visual token pruning for multi-modal large language models},
  author={Ye, Weihao and Wu, Qiong and Lin, Wenhao and Zhou, Yiyi},
  booktitle={Proceedings of the AAAI Conference on Artificial Intelligence},
  pages={22128--22136},
  year={2025}
}

@article{pyramiddrop,
  title={Pyramiddrop: Accelerating your large vision-language models via pyramid visual redundancy reduction},
  author={Xing, Long and Huang, Qidong and Dong, Xiaoyi and Lu, Jiajie and Zhang, Pan and Zang, Yuhang and Cao, Yuhang and He, Conghui and Wang, Jiaqi and Wu, Feng and others},
  journal={arXiv preprint arXiv:2410.17247},
  year={2024}
}

@inproceedings{sparsevila,
  title={SparseVILA: Decoupling Visual Sparsity for Efficient VLM Inference},
  author={Khaki, Samir and Guo, Junxian and Tang, Jiaming and Yang, Shang and Chen, Yukang and Plataniotis, Konstantinos N and Lu, Yao and Han, Song and Liu, Zhijian},
  booktitle={Proceedings of the IEEE/CVF International Conference on Computer Vision},
  pages={23784--23794},
  year={2025}
}

@misc{dao2022flashattentionfastmemoryefficientexact,
      title={FlashAttention: Fast and Memory-Efficient Exact Attention with IO-Awareness}, 
      author={Tri Dao and Daniel Y. Fu and Stefano Ermon and Atri Rudra and Christopher Ré},
      year={2022},
      eprint={2205.14135},
      archivePrefix={arXiv},
      primaryClass={cs.LG},
      url={https://arxiv.org/abs/2205.14135}, 
}

@article{sparsevlm,
  title={Sparsevlm: Visual token sparsification for efficient vision-language model inference},
  author={Zhang, Yuan and Fan, Chun-Kai and Ma, Junpeng and Zheng, Wenzhao and Huang, Tao and Cheng, Kuan and Gudovskiy, Denis and Okuno, Tomoyuki and Nakata, Yohei and Keutzer, Kurt and others},
  journal={arXiv preprint arXiv:2410.04417},
  year={2024}
}

@article{fastervlm,
  title={[CLS] Attention is All You Need for Training-Free Visual Token Pruning: Make VLM Inference Faster},
  author={Zhang, Qizhe and Cheng, Aosong and Lu, Ming and Zhuo, Zhiyong and Wang, Minqi and Cao, Jiajun and Guo, Shaobo and She, Qi and Zhang, Shanghang},
  journal={arXiv e-prints},
  pages={arXiv--2412},
  year={2024}
}

@inproceedings{hired,
  title={HiRED: Attention-Guided Token Dropping for Efficient Inference of High-Resolution Vision-Language Models},
  author={Arif, Kazi Hasan Ibn and Yoon, JinYi and Nikolopoulos, Dimitrios S and Vandierendonck, Hans and John, Deepu and Ji, Bo},
  booktitle={Proceedings of the AAAI Conference on Artificial Intelligence},
  volume={39},

  pages={1773--1781},
  year={2025}
}

@InProceedings{visionzip,
    author    = {Yang, Senqiao and Chen, Yukang and Tian, Zhuotao and Wang, Chengyao and Li, Jingyao and Yu, Bei and Jia, Jiaya},
    title     = {VisionZip: Longer is Better but Not Necessary in Vision Language Models},
    booktitle = {Proceedings of the IEEE/CVF Conference on Computer Vision and Pattern Recognition (CVPR)},
    month     = {June},
    year      = {2025},
    pages     = {19792-19802}
}

@article{holov,
  title={Don't Just Chase" Highlighted Tokens" in MLLMs: Revisiting Visual Holistic Context Retention},
  author={Zou, Xin and Lu, Di and Wang, Yizhou and Yan, Yibo and Lyu, Yuanhuiyi and Zheng, Xu and Zhang, Linfeng and Hu, Xuming},
  journal={arXiv preprint arXiv:2510.02912},
  year={2025}
}

@article{cdpruner,
  title={Beyond Attention or Similarity: Maximizing Conditional Diversity for Token Pruning in MLLMs},
  author={Zhang, Qizhe and Liu, Mengzhen and Li, Lichen and Lu, Ming and Zhang, Yuan and Pan, Junwen and She, Qi and Zhang, Shanghang},
  journal={arXiv preprint arXiv:2506.10967},
  year={2025}
}

@inproceedings{divprune,
  title={Divprune: Diversity-based visual token pruning for large multimodal models},
  author={Alvar, Saeed Ranjbar and Singh, Gursimran and Akbari, Mohammad and Zhang, Yong},
  booktitle={Proceedings of the Computer Vision and Pattern Recognition Conference},
  pages={9392--9401},
  year={2025}
}

@inproceedings{vispruner,
  title={Beyond text-visual attention: Exploiting visual cues for effective token pruning in vlms},
  author={Zhang, Qizhe and Cheng, Aosong and Lu, Ming and Zhang, Renrui and Zhuo, Zhiyong and Cao, Jiajun and Guo, Shaobo and She, Qi and Zhang, Shanghang},
  booktitle={Proceedings of the IEEE/CVF International Conference on Computer Vision},
  pages={20857--20867},
  year={2025}
}

@article{dart,
  title={Stop looking for important tokens in multimodal language models: Duplication matters more},
  author={Wen, Zichen and Gao, Yifeng and Wang, Shaobo and Zhang, Junyuan and Zhang, Qintong and Li, Weijia and He, Conghui and Zhang, Linfeng},
  journal={arXiv preprint arXiv:2502.11494},
  year={2025}
}

@inproceedings{vqav2,
  title={Making the v in vqa matter: Elevating the role of image understanding in visual question answering},
  author={Goyal, Yash and Khot, Tejas and Summers-Stay, Douglas and Batra, Dhruv and Parikh, Devi},
  booktitle={Proceedings of the IEEE conference on computer vision and pattern recognition},
  pages={6904--6913},
  year={2017}
}

@inproceedings{gqa,
  title={Gqa: A new dataset for real-world visual reasoning and compositional question answering},
  author={Hudson, Drew A and Manning, Christopher D},
  booktitle={Proceedings of the IEEE/CVF conference on computer vision and pattern recognition},
  pages={6700--6709},
  year={2019}
}

@article{sqa,
  title={Learn to explain: Multimodal reasoning via thought chains for science question answering},
  author={Lu, Pan and Mishra, Swaroop and Xia, Tanglin and Qiu, Liang and Chang, Kai-Wei and Zhu, Song-Chun and Tafjord, Oyvind and Clark, Peter and Kalyan, Ashwin},
  journal={Advances in Neural Information Processing Systems},
  volume={35},
  pages={2507--2521},
  year={2022}
}

@inproceedings{textvqa,
  title={Towards vqa models that can read},
  author={Singh, Amanpreet and Natarajan, Vivek and Shah, Meet and Jiang, Yu and Chen, Xinlei and Batra, Dhruv and Parikh, Devi and Rohrbach, Marcus},
  booktitle={Proceedings of the IEEE/CVF conference on computer vision and pattern recognition},
  pages={8317--8326},
  year={2019}
}

@article{pope,
  title={Evaluating object hallucination in large vision-language models},
  author={Li, Yifan and Du, Yifan and Zhou, Kun and Wang, Jinpeng and Zhao, Wayne Xin and Wen, Ji-Rong},
  journal={arXiv preprint arXiv:2305.10355},
  year={2023}
}

@inproceedings{mme,
  title={Mme: A comprehensive evaluation benchmark for multimodal large language models},
  author={Fu, Chaoyou and Chen, Peixian and Shen, Yunhang and Qin, Yulei and Zhang, Mengdan and Lin, Xu and Yang, Jinrui and Zheng, Xiawu and Li, Ke and Sun, Xing and others},
  booktitle={The Thirty-ninth Annual Conference on Neural Information Processing Systems Datasets and Benchmarks Track},
  year={2025}
}

@inproceedings{mmb,
  title={Mmbench: Is your multi-modal model an all-around player?},
  author={Liu, Yuan and Duan, Haodong and Zhang, Yuanhan and Li, Bo and Zhang, Songyang and Zhao, Wangbo and Yuan, Yike and Wang, Jiaqi and He, Conghui and Liu, Ziwei and others},
  booktitle={European conference on computer vision},
  pages={216--233},
  year={2024},
  organization={Springer}
}

@article{mmvet,
  title={Mm-vet: Evaluating large multimodal models for integrated capabilities},
  author={Yu, Weihao and Yang, Zhengyuan and Li, Linjie and Wang, Jianfeng and Lin, Kevin and Liu, Zicheng and Wang, Xinchao and Wang, Lijuan},
  journal={arXiv preprint arXiv:2308.02490},
  year={2023}
}

@misc{zhou2025mlvubenchmarkingmultitasklong,
      title={MLVU: Benchmarking Multi-task Long Video Understanding}, 
      author={Junjie Zhou and Yan Shu and Bo Zhao and Boya Wu and Zhengyang Liang and Shitao Xiao and Minghao Qin and Xi Yang and Yongping Xiong and Bo Zhang and Tiejun Huang and Zheng Liu},
      year={2025},
      eprint={2406.04264},
      archivePrefix={arXiv},
      primaryClass={cs.CV},
      url={https://arxiv.org/abs/2406.04264}, 
}

@misc{li2024mvbenchcomprehensivemultimodalvideo,
      title={MVBench: A Comprehensive Multi-modal Video Understanding Benchmark}, 
      author={Kunchang Li and Yali Wang and Yinan He and Yizhuo Li and Yi Wang and Yi Liu and Zun Wang and Jilan Xu and Guo Chen and Ping Luo and Limin Wang and Yu Qiao},
      year={2024},
      eprint={2311.17005},
      archivePrefix={arXiv},
      primaryClass={cs.CV},
      url={https://arxiv.org/abs/2311.17005}, 
}

@misc{fu2025videommefirstevercomprehensiveevaluation,
      title={Video-MME: The First-Ever Comprehensive Evaluation Benchmark of Multi-modal LLMs in Video Analysis}, 
      author={Chaoyou Fu and Yuhan Dai and Yongdong Luo and Lei Li and Shuhuai Ren and Renrui Zhang and Zihan Wang and Chenyu Zhou and Yunhang Shen and Mengdan Zhang and Peixian Chen and Yanwei Li and Shaohui Lin and Sirui Zhao and Ke Li and Tong Xu and Xiawu Zheng and Enhong Chen and Caifeng Shan and Ran He and Xing Sun},
      year={2025},
      eprint={2405.21075},
      archivePrefix={arXiv},
      primaryClass={cs.CV},
      url={https://arxiv.org/abs/2405.21075}, 
}

@misc{zhao2025mmvu,
      title={MMVU: Measuring Expert-Level Multi-Discipline Video Understanding}, 
      author={Yilun Zhao and Lujing Xie and Haowei Zhang and Guo Gan and Yitao Long and Zhiyuan Hu and Tongyan Hu and Weiyuan Chen and Chuhan Li and Junyang Song and Zhijian Xu and Chengye Wang and Weifeng Pan and Ziyao Shangguan and Xiangru Tang and Zhenwen Liang and Yixin Liu and Chen Zhao and Arman Cohan},
      year={2025},
      eprint={2501.12380},
      archivePrefix={arXiv},
      primaryClass={cs.CV},
      url={https://arxiv.org/abs/2501.12380}, 
}

@misc{he2024mmworld,
      title={MMWorld: Towards Multi-discipline Multi-faceted World Model Evaluation in Videos}, 
      author={Xuehai He and Weixi Feng and Kaizhi Zheng and Yujie Lu and Wanrong Zhu and Jiachen Li and Yue Fan and Jianfeng Wang and Linjie Li and Zhengyuan Yang and Kevin Lin and William Yang Wang and Lijuan Wang and Xin Eric Wang},
      year={2024},
      eprint={2406.08407},
      archivePrefix={arXiv},
      primaryClass={cs.CV},
      url={https://arxiv.org/abs/2406.08407}, 
}

@article{yarbuseye,
author = {Benjamin W Tatler;Nicholas J Wade;Hoi Kwan;John M Findlay;Boris M Velichkovsky;},
title = {Yarbus, Eye Movements, and Vision},
journal = {i-Perception},
volume = {1},
number = {1},
pages = {7-27},
year = {2010},
doi = {10.1068/i0382},
URL = {http://dx.doi.org/10.1068/i0382},
eprint = {http://dx.doi.org/10.1068/i0382} 
}

@article{cogitiveding,
author = {Ding, Wenjun and Yu, Guoxing},
year = {2025},
month = {12},
pages = {1-30},
title = {Young Learners’ Cognitive Processes in Picture-Based Causal Explanation Speaking Tasks: Synchronizing Eye Movements with Speech Production},
volume = {22},
journal = {Language Assessment Quarterly},
doi = {10.1080/15434303.2025.2604719}
}

@article{henderson2003human,
title = {Human gaze control during real-world scene perception},
journal = {Trends in Cognitive Sciences},
volume = {7},
number = {11},
pages = {498-504},
year = {2003},
issn = {1364-6613},
doi = {https://doi.org/10.1016/j.tics.2003.09.006},
url = {https://www.sciencedirect.com/science/article/pii/S1364661303002481},
author = {John M. Henderson}
}

@article{wolfe2017five,
author = {Wolfe, Jeremy and Horowitz, Todd},
year = {2017},
month = {03},
pages = {0058},
title = {Five Factors that Guide Attention in Visual Search},
volume = {1},
journal = {Nature Human Behaviour},
doi = {10.1038/s41562-017-0058}
}

@article{alvarez2011representing,
author = {Alvarez, George},
year = {2011},
month = {02},
pages = {122-31},
title = {Representing multiple objects as an ensamble enhance visual cognition},
volume = {15},
journal = {Trends in cognitive sciences},
doi = {10.1016/j.tics.2011.01.003}
}

@article{Gonzalez1985ClusteringTM,
  title={Clustering to Minimize the Maximum Intercluster Distance},
  author={Teofilo F. Gonzalez},
  journal={Theor. Comput. Sci.},
  year={1985},
  volume={38},
  pages={293-306},
  url={https://api.semanticscholar.org/CorpusID:205092276}
}

@article{hochbaum1985best,
 ISSN = {0364765X, 15265471},
 URL = {http://www.jstor.org/stable/3689371},
 author = {Dorit S. Hochbaum and David B. Shmoys},
 journal = {Mathematics of Operations Research},
 number = {2},
 pages = {180--184},
 publisher = {INFORMS},
 title = {A Best Possible Heuristic for the k-Center Problem},
 urldate = {2026-01-25},
 volume = {10},
 year = {1985}
}

\end{document}